%% file: main.tex
\documentclass{article} 
\usepackage{iclr2024_conference,times}

\input{math_commands.tex}

\usepackage{hyperref}
\usepackage{url}

\usepackage{graphicx}
\usepackage{multirow}
\usepackage{booktabs}
\usepackage{colortbl}   
\usepackage{xcolor}
\usepackage{tikz}

\usepackage{wrapfig}
\usepackage{tabularx}
\usepackage{subfigure}
\usepackage{caption}
\usepackage[most]{tcolorbox}
\newtcolorbox{response}[1][]{
  colback=blue!2,
  colframe=black!75,
  fonttitle=\bfseries,
  coltitle=black,
  }
\usepackage{marvosym}

\title{Federated Learning Empowered by \\Generative Content}

\author{Rui Ye\textsuperscript{1}, Xinyu Zhu\textsuperscript{1}, Jingyi Chai\textsuperscript{1}, Siheng Chen\textsuperscript{1,2\Letter}, Yanfeng Wang\textsuperscript{2,1} \\
\textsuperscript{1}Shanghai Jiao Tong University, \textsuperscript{2}Shanghai AI Laboratory \\
\texttt{\{yr991129,zhuxinyu,chaijingyi,sihengc,wangyanfeng\}@sjtu.edu.cn} \vspace{-5mm}
}

\iclrfinalcopy 
\begin{document}

\maketitle

\begin{abstract}

Federated learning (FL) enables leveraging distributed private data for model training in a privacy-preserving way. 
However, data heterogeneity significantly limits the performance of current FL methods. 
In this paper, we propose a novel FL framework termed FedGC, designed to mitigate data heterogeneity issues by diversifying private data with generative content.
FedGC is a simple-to-implement framework as it only introduces a one-shot step of data generation.
In data generation, we summarize three crucial and worth-exploring aspects (budget allocation, prompt design, and generation guidance) and propose three solution candidates for each aspect.
Specifically, to achieve a better trade-off between data diversity and fidelity for generation guidance, we propose to generate data based on the guidance of prompts and real data simultaneously.
The generated data is then merged with private data to facilitate local model training.
Such generative data increases the diversity of private data to prevent each client from fitting the potentially biased private data, alleviating the issue of data heterogeneity.
We conduct a systematic empirical study on FedGC, covering diverse baselines, datasets, scenarios, and modalities.
Interesting findings include (1) FedGC consistently and significantly enhances the performance of FL methods, even when notable disparities exist between generative and private data; 
(2) FedGC achieves both better performance and privacy-preservation.
We wish this work can inspire future works to further explore the potential of enhancing FL with generative content.

\end{abstract}

\section{Introduction}

Federated learning (FL) is a privacy-preserving machine learning paradigm that enables multiple clients to collaboratively train a global model without directly sharing their raw data~\citep{fedavg,advances}. With the increasing concerns about privacy, FL has attracted significant attention and has been applied to diverse real-world fields such as natural language processing, healthcare, finance, Internet of Things (IoT), and autonomous vehicles~\citep{yang_survey}.

Data heterogeneity presents a prominent and fundamental challenge in FL, significantly impacting FL's overall performance~\citep{fedavg,fedavgm}. This heterogeneity arises inherently due to the varied environments and preferences in which clients' datasets are collected. Consequently, it results in biased and divergent local models, posing difficulties in achieving a well-generalized aggregated model capable of effectively addressing diverse data sources.

Addressing this issue, many optimization-based works are proposed from diverse perspectives~\citep{wangsurvey}. 
On the client side, they regularize the distance between local and global model~\citep{fedprox,feddyn}, introduce control variates to correct local gradients~\citep{scaffold}, align the feature space~\citep{fedfm,moon}. 
On the server side, they introduce momentum to update global model~\citep{fedopt,fedavgm}, adjust the process of aggregating local models~\citep{fednova,fedexp}, modify model initialization~\citep{nguyen2022begin,chen2022importance}.
However, the performance of all these methods is still severely limited as data heterogeneity fundamentally exists.

In this paper, we propose a new idea of fundamentally mitigating the effects of data heterogeneity with the help of diverse generative content.
To realize this idea, we propose a novel framework, Federated Learning with Generative Content (FedGC).
In FedGC, each client uses a publicly available generative model conditioned on task-related prompts to generate diverse data, which supplements the originally client-specific (the root of data heterogeneity) data.
The supplemented dataset can subsequently facilitate client model training by encouraging the local model to also learn diverse patterns rather than only patterns of its private data.
Despite the simplicity, FedGC can significantly mitigate data heterogeneity as generative diverse data introduces informative and general patterns, thus preventing each client from over-fitting its potentially biased private data.

Furthermore, FedGC is a flexible framework with multiple potential directions.
Considering generation efficiency, data diversity, and data fidelity, we summarize four critical aspects in FedGC, including budget allocation, prompt design, generation guidance, and training strategy.
In each aspect, we propose three representative solutions as candidates.
For example, to achieve a better trade-off between diversity and fidelity during generation, we propose real-data-guidance which generates data conditioned on real data and task-related prompts simultaneously.

To prove the effectiveness of FedGC and deepen understanding, we conduct a systematic empirical study from diverse perspectives, including compatibility with FL baselines, different datasets, different modalities, and different data heterogeneity types; and have several interesting findings.
1) Adding generative data is a more direct, concise, and effective solution to tackle data heterogeneity, than many sophisticated algorithm designs. 
2) FedGC can achieve both better privacy preservation and performance.
3) Despite failing to resemble real data, generative data still contributes to enhanced performance as it can implicitly reduce data heterogeneity and model divergence.

Our contributions are as follows:
\begin{enumerate}
    \item We propose FedGC, a new, simple yet effective FL framework that handles data heterogeneity from a new perspective: generating diverse data to supplement private real data.
    \item We summarize four critical and worth-exploring facets in FedGC and propose three solution candidates for each, underscoring its flexibility and potential for future explorations.
    \item We provide a systematic empirical study on FedGC framework, showing its effectiveness for tackling data heterogeneity and providing new insights for future works through several interesting experimental findings.
\end{enumerate}

\section{Related Work}

\textbf{Federated learning} (FL) enables multiple clients to collaboratively train a global model without sharing raw data~\citep{fedavg}, which has attracted much attention due to its privacy-preserving property~\citep{li_survey,advances}. 
Data heterogeneity is one representative challenge in FL that significantly limits the FL's performance~\citep{fedavgm,li2019convergence}. 
Addressing this, many methods are proposed to mitigate its adverse effects from the perspective of optimization. 
(1) On client-side optimization, FedProx~\citep{fedprox} and SCAFFOLD~\citep{scaffold} propose to conduct model-level correction such as regularizing $\ell_2$ distance between local and global model and introducing a control variate to correct gradient of local model. 
MOON~\citep{moon} and FedDecorr~\citep{feddecorr} propose to regularize feature space.
(2) On server-side optimization, FedNova~\citep{fednova} and FedDisco~\citep{feddisco} propose to modify aggregation weights to obtain better-aggregated model.
\citep{nguyen2022begin,chen2022importance} explore the effects of model initialization.
FedAvgM~\citep{fedavgm} and FedOPT~\citep{fedopt} apply momentum-based optimization to improve global model updating.

Unlike these optimization-level methods that still fundamentally suffer from data heterogeneity, our FedGC framework focuses on data-level improvement, which mitigates heterogeneity of the distributed real data by complementing it with diverse generative data. 
Besides, our FedGC framework is orthogonal to these methods, allowing seamless integration within our framework.

\textbf{Generative models} have demonstrated remarkable performance across multiple domains such as large language models~\citep{ouyang2022training,openai2023gpt4,llama2} for language generation and diffusion models~\citep{nichol2022glide,rombach2022high,saharia2022photorealistic} for image generation. 
Though these models can generate high-quality data for general cases, the generated data is not sufficient to train a well-perform model due to its incapability of representing real data~\citep{he2022synthetic}, especially for uncommon cases such as medical tasks~\citep{eysenbach2023role,celard2023survey}.
Recently, \citep{shipard2023diversity} shows the importance of data diversity for zero-shot image classification tasks.

In this paper, we systematically explore the potential of using generative models to assist FL on private data. Based on our FedGC framework, we verify that despite failing to fully represent real data, generated data can still contribute to improving the performance of FL under heterogeneous private data. Besides, FedGC is applicable to both image and text regimes.

\section{Federated Learning with Generative Content}

We propose FedGC, a new FL framework that leverages generative content to tackle the issue of data heterogeneity in FL. 
Based on FedGC, we summarize four aspects worth exploring and propose three methods for each aspect, which serves to provide more insights for future works. 

\subsection{FedGC Framework Overview}

\begin{figure}[t]
\begin{center}
\includegraphics[width=0.9\textwidth]{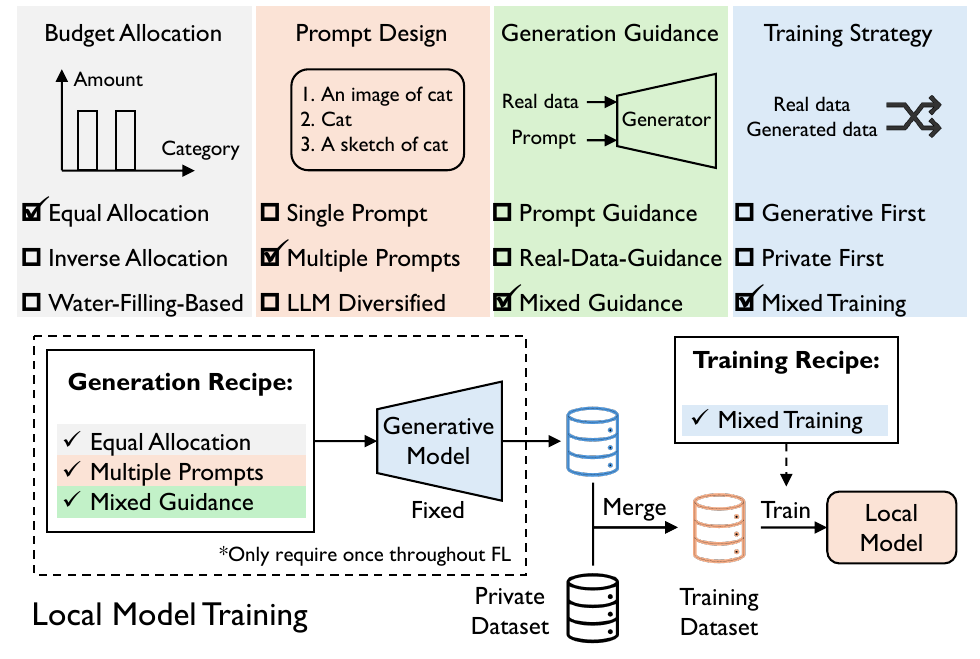}
\end{center}
\caption{Overview of the designs of FedGC on client side. Above, we summarize four crucial aspects that are worth exploring and propose three solutions for each aspect. Below is the pipeline of local training, where each client first generates data based on the generation recipe, then merges the generative and private dataset, and finally trains the local model based on the training recipe.}
\label{fig:overview}
\end{figure}

Our FedGC follows the standard FedAvg framework, encompassing of four iterative phases: global model broadcasting, local model training, local model uploading, and global model aggregation.
Our goal is to generate diverse data to supplement private data to facilitate local model training.
Considering communication cost and flexibility, we generate data on the client (local) side, which avoids additional communication cost required for server-to-client transmitting generative data, and enables using the local data as prior to generate more specific data.
Thus, we focus on local model training, which is decomposed into: \underline{data generation} and \underline{local model training}. 
Specifically, in FedGC, we 1) design to generate diverse data, 2) merge the generative and private dataset, and 3) train the local model, where the first two are required for only once; see Figure~\ref{fig:overview} for the overview. Although FedGC is versatile across modalities, our illustration herein will on images for easier understanding.

\subsection{Data Generation in FedGC}
 
On the designs for data generation in FedGC framework, we consider the following criteria: \underline{generation efficiency}, \underline{data diversity}, and \underline{data fidelity}.
Following the criteria, we explore three crucial aspects, including budget allocation, prompt design, and generation guidance, and propose three representative solutions as candidates for each aspect.
Without loss of generality, we use the text-guided latent diffusion model~\citep{rombach2022high} to generate images based on prompts for image task, and ChatGPT~\cite{openai2023gpt4} to generate texts based on prompts for text task.

\textbf{Budget allocation for efficiency.} 
Though, (1) the process of data generation is just one-shot and (2) FedGC does not compromise on the two first-order concerns in FL: communication cost and privacy~\citep{advances}, it still costs some computation budget in exchange for algorithm utility~\citep{no_free_lunch}. 
Thus, it is essential to design efficient strategies to allocate the generation budget (i.e., the total number of generative samples, denoted as $M$) to each client and label. 

To achieve this, we design three allocation strategies. 
(1) The equal allocation strategy allocates the budget equally to each client and each category, which is the simplest and most general allocation strategy. 
That is, each client can generate $\frac{M}{KC}$ data samples for each category.
(2) Inverse allocation strategy allocates the budget inversely to each client according its number of data samples. 
Specifically, each client $k$ can generate $\frac{M \cdot (N_{max}-N_k)}{C \cdot \sum_i (N_{max}-N_i)}$ samples for each category, where $N_{max}$ denotes the maximum number in $\{N_i\}_i$. 
(3) Water-filling-based: each client can generate $\frac{M}{K}$ samples in total, and apply water filling algorithm to allocate samples to each category~\citep{proakis2008digital}.

\textbf{Prompt design for diversity.}
Data diversity plays a key role in learning a generalized model in many domains such as image~\citep{chen2020simple} and text~\citep{radford2019language}.
To increase the diversity, it is essential to design appropriate prompts since they directly guide the process of generation.

For image task, we consider three diversity levels.
(1) Single prompt, where we use ``a photo of \{class\}''~\citep{radford2021learning}.
(2) Multiple prompts, where we consider diverse formats such as ``\{class\}''.
(3) LLM-based diversified prompts, where we instruct an LLM such as ChatGPT to diversify the prompts.
While for text generation, we only design one prompt since the ChatGPT~\citep{openai2023gpt4} is sufficient to generate diverse content if we instruct it to be diverse; see Table~\ref{tab:real-data-guide-text}.

\textbf{Generation guidance for diversity and fidelity.}
Finally, we feed the prompts to the generative models for generation. 
Besides designing prompts, we randomly set the guidance scale for diffusion models~\citep{rombach2022high} (or non-zero temperature for LLMs) to enhance the data diversity.

(Prompt-Only Guidance) However, data diversity may not be sufficient to ensure improving model training, while data fidelity is also a critical factor. 
For cases where the domain gap between the generative and real data is too large, the benefits of increasing diversity may be outweighed by the negative effects of the domain gap, leading to degraded performance~\citep{he2022synthetic}.

(Real-Data Guidance) To alleviate this issue, we propose a new real-data-guided generation approach, which conditions data generation on both real data and prompts. 
For image task, unlike the original text-guided generation that starts from a random Gaussian noise at latent space $\vz^1_T$~\citep{rombach2022high}, we propose to inject information of real data into the starting noise. 
Specifically, we first use the auto-encoder to encode the real image $\vx$ to latent representation $\vz$, then add some Gaussian variation to obtain a new $\vz^2_T$, which substitutes $\vz^1_T$ as the starting point; see illustration in ~\ref{fig:real-data-guide-img}.
This enriched latent representation, infused with real data insights, enables the generative model to produce outputs closely resembling real data, optimizing the trade-off between diversity and fidelity. 
For text task, see illustration in Table~\ref{tab:real-data-guide-text} using ChatGPT.

(Mixed Guidance) Furthermore, given that certain clients may lack data samples from specific categories, we propose a mixed guidance strategy.
Specifically, for a given budget $N_{k,c}$ for client $k$ in category $c$,
(1) if client $k$ possesses samples from category $c$, it generates $N_{k,c}/2$ samples using text-only guidance and $N_{k,c}/2$ samples with real-data guidance; 
(2) in the absence of samples for client $k$ from category $c$, it generates all the $N_{k,c}$ samples using text-only guidance. 
This approach effectively addresses category omissions and refines the trade-off between diversity and fidelity.

\subsection{Local Model Training in FedGC}

By choosing generation recipe from the three aspects above, we can generate data using the generative model to assist local model training. 
Given the generative dataset $\mathcal{D}_g$ and the private dataset $\mathcal{D}_p$, there could be diverse training strategies such as sequential training (optimizing on the two datasets sequentially) and mixed training (optimizing on the mixed dataset).

We find that the mixed training strategy is the most effective despite its simplicity.
Thus, we directly merge the two datasets as the final new training dataset $\mathcal{D}_m$, based on which we train the local model at the same training manner protocol as other FL methods. Specifically, at the $t$-th FL communication round, each client $k$ first receives the global model $\bm{\theta}^t$ and re-initializes its local model with $\bm{\theta}^t$. Then, each client conducts model training based on the merged dataset $\mathcal{D}_m$ for several optimization steps. Finally, each client $k$ obtains its local model $\bm{\theta}_k^t$, which is subsequently sent to the server for model aggregation ($\bm{\theta}^{t+1} := \sum_k p_k \bm{\theta}_k^t$, where $p_k=N_k/\sum_i N_i$ is the relative dataset size).

Note that this process is orthogonal to local training algorithm, which can be SGD-based training~\citep{fedavg}, proximity-based training~\citep{fedprox}, control-variate-based training~\citep{scaffold}, or feature-alignment-based training~\citep{moon}.

\section{Experiments}

\begin{table}[t]
\setlength\tabcolsep{4.5pt}
\caption{Experiments on two heterogeneity types, four datasets, two heterogeneity levels, and six baselines. Test accuracy (\%) averaged over three trials is reported. FedGC consistently and significantly brings performance gain over baselines across diverse settings.}
\label{tab:main}
\begin{center}
\begin{tabular}{cc|cccc|cccc|c}
\toprule
\multirow{3}{*}{Baseline} & H-Type & \multicolumn{4}{c|}{Label Level} & \multicolumn{4}{c|}{Feature Level} & \textbf{Avg.}\\
& Dataset & \multicolumn{2}{c}{CIFAR-10}  & \multicolumn{2}{c|}{EuroSAT} & \multicolumn{2}{c}{PACS} & \multicolumn{2}{c|}{VLCS} & \textbf{Acc.}\\
& H-Level & High  & Low  & High  & Low & High  & Low & High  & Low & $\bm{\Delta}$\\
\midrule
 & Vanilla & 61.25 & 75.88 &  53.82 & 75.59 & 27.16 & 36.47 & 43.69 & 47.95 & \multirow{2}{*}{\textbf{+12.26}} \\
\rowcolor{gray!15} \multirow{-2}{*}{\cellcolor{white}FedAvg} & + FedGC & 74.50 & 79.73 & 74.83 & 84.46 & 54.43 & 53.93 & 46.49 & 50.50 \\ 
& Vanilla & 60.83 & 74.40 & 50.91 & 72.80 & 28.96 & 34.52 & 46.64 & 45.74 & \multirow{2}{*}{\textbf{+13.04}} \\
\rowcolor{gray!15} \multirow{-2}{*}{\cellcolor{white}FedAvgM} & + FedGC & 73.84 & 78.90 & 73.48 & 84.87 & 53.23 & 55.73 & 48.45 & 50.65 \\
& Vanilla & 64.02 & 75.62 & 59.61 & 73.20 & 27.71 & 39.52 & 38.83 & 48.50 & \multirow{2}{*}{\textbf{+11.49}}\\
\rowcolor{gray!15} \multirow{-2}{*}{\cellcolor{white}FedProx} &+ FedGC & 74.36 & 79.25  & 73.04 & 84.76 & 54.28 & 55.83 & 45.69 & 51.70  \\ 
& Vanilla & 63.98 & 78.79 & 52.72 & 76.80 & 29.72 & 37.52 & 43.64 & 40.83 & \multirow{2}{*}{\textbf{+12.57}}\\
\rowcolor{gray!15} \multirow{-2}{*}{\cellcolor{white}SCAFFOLD} &+ FedGC & 73.96 & 80.29 & 69.48 & 81.04 & 59.73 & 60.63 & 47.65 & 51.75  \\ 
& Vanilla & 63.40 & 75.43 & 52.67 & 70.02 & 27.91 & 36.52 & 45.89 & 48.30 & \multirow{2}{*}{\textbf{+12.47}}\\
\rowcolor{gray!15} \multirow{-2}{*}{\cellcolor{white}MOON} &+ FedGC & 74.02 & 79.82 & 73.69 & 86.06 & 53.81 & 55.08 & 48.05 & 49.35  \\ 
& Vanilla & 64.14 & 76.19 & 63.74 & 69.57 & 27.51 & 29.07 & 37.02 & 47.70 & \multirow{2}{*}{\textbf{+9.62}}\\
\rowcolor{gray!15} \multirow{-2}{*}{\cellcolor{white}FedDecorr} &+ FedGC & 73.94 & 78.16 & 69.93 & 81.30 & 48.77 & 47.42 & 43.39 & 49.00 \\
\bottomrule
\end{tabular}
\end{center}
\end{table}

\subsection{Implementation Details}

We set the number of communication rounds as 100. Table~\ref{tab:client_number} lists client number for each dataset.

\textbf{Data Heterogeneity and Datasets.}
We consider two types of data heterogeneity for image tasks. 
For label heterogeneity, we consider CIFAR-10~\citep{cifar10} and EuroSAT~\citep{helber2019eurosat}, where we allocate the original training dataset to clients based on the frequently used strategy in FL: Dirichlet distribution~\citep{fedma}. 
$\beta$ controls the level of heterogeneity, where we denote $0.05$ as high and $0.1$ as low. 
For feature heterogeneity, we consider PACS~\citep{pacs} and VLCS~\citep{vlcs}, where we allocate training dataset of each domain to several clients according to Dirichlet distribution.
This captures both the properties of feature- and label-level heterogeneity.
For text datasets, we consider Sentiment140 from LEAF benchmark~\citep{leaf} (naturally allocated) and Yahoo! Answers~\citep{yahoo} (split by Dirichlet distribution).

\textbf{Training Details.} 
The number of iterations for local model training is 200 and uses SGD as the optimizer with a batch size of 64.
The learning rate is set to 0.01~\citep{moon,feddisco}.
We use ResNet-20~\citep{resnet} for image task and LSTM for text task~\citep{leaf}.

\subsection{Experimental Results}

\textbf{FedGC significantly improves the FL performance under data heterogeneity.} In Table~\ref{tab:main}, we show experimental results on two heterogeneity types (label-level and feature-level heterogeneity), two datasets for each type (CIFAR-10, EuroSAT, PACS, and VLCS), and two heterogeneity levels for each dataset.
From the table, we see that (1) incorporating baseline in our FedGC framework can consistently and significantly improve the performance of baseline across diverse settings.
(2) FedGC is extremely helpful when the heterogeneity level is relatively high, convincingly supporting our idea of introducing generative data to mitigate the effects of data heterogeneity. 
Specifically, based on FedAvg, FedGC brings 21.01 absolute accuracy improvement under a high heterogeneity level on EuroSAT and 12.26 absolute accuracy improvement on average.

\textbf{FedGC is compatible with existing FL methods.} From Table~\ref{tab:main}, we see that FedGC consistently and significantly brings performance gain across 6 different baselines, including FedAvg, FedAvgM, FedProx, SCAFFOLD, MOON, and FedDecorr. 
For example, FedGC averagely brings 12.68 absolute accuracy improvement to SCAFFOLD~\citep{scaffold}.
This demonstrates the compatibility and universality of our proposed FedGC framework.
 
\begin{wrapfigure}{r}{7.2cm}
\vspace{-15pt}
   \centering
    \subfigure[50k Real Samples]{
		\includegraphics[width=0.24\columnwidth]{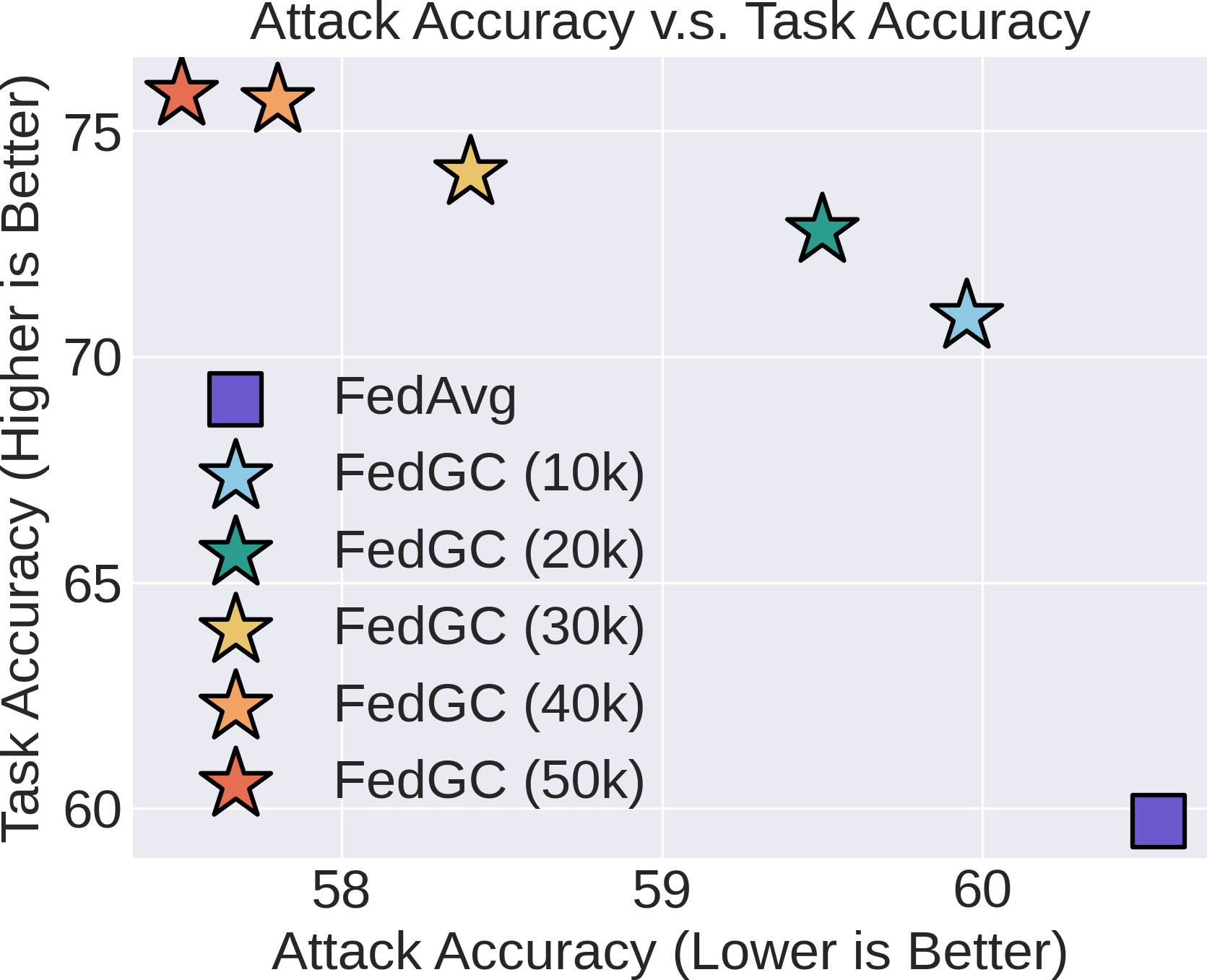}
		\label{fig:5w}
	}
	\subfigure[10k Real Samples]{
		\includegraphics[width=0.24\columnwidth]{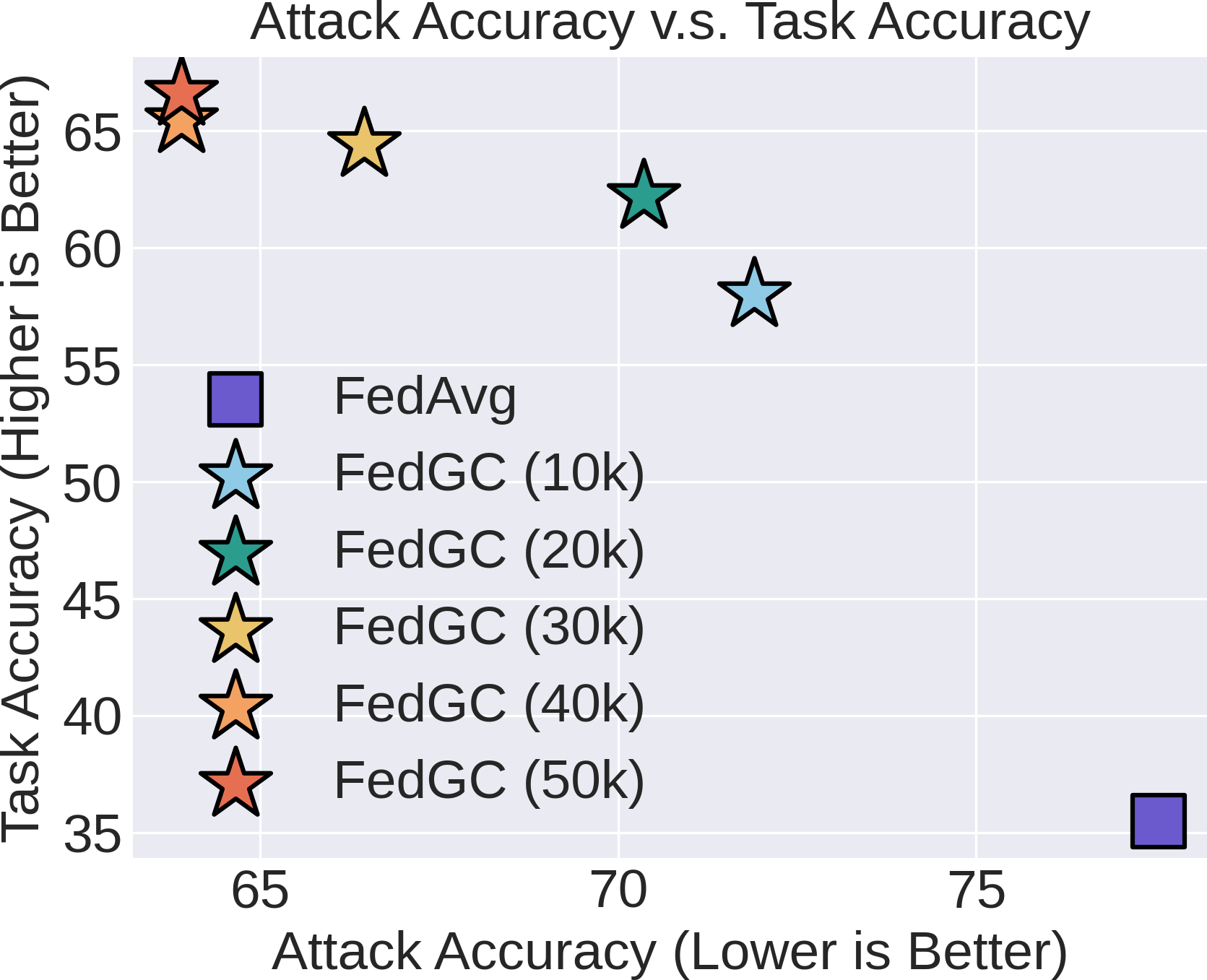}
		\label{fig:1w}
	}
    \vspace{-10pt}
   \caption{FedGC achieves both better task accuracy and privacy preservation (lower attack accuracy).}
   \vspace{-10pt}
   \label{fig:privacy_acc}
\end{wrapfigure}

\textbf{FedGC achieves better performance and privacy preservation at the same time.} In Figure~\ref{fig:privacy_acc}, we show the performance and privacy preservation trade-off comparisons before and after using FedGC.
To measure privacy preservation, we use a simple membership inference attack method based on loss evaluation~\citep{yu2021does,sablayrolles2019white} to evaluate attack accuracy, see details in Section~\ref{app:mia}.
Lower attack accuracy indicates better privacy preservation.
From the figure, we have an interesting finding that our FedGC framework can not only improve the performance under data heterogeneity, but also enhance the privacy preservation.
We also show in Table~\ref{tab:mia} that FedGC achieves significantly lower attack accuracy when similar task accuracy is achieved.
This is surprising yet reasonable since FedGC requires the model to learn from both the private data and the diverse generative data, meaning that the generative data can dilute the concentration of real, sensitive data.

This explanation can be further verified since (1) as the number of generated samples increases, FedGC achieves lower attack accuracy (better privacy preservation). (2) When the number of real training samples is smaller (from 50k to 10k), we see a much larger reduction in attack accuracy and improvement in task accuracy, since the ratio of private data samples in the whole dataset is lowered.

\begin{wrapfigure}{r}{7cm}
\vspace{-15pt}
   \centering
    \subfigure[Sentiment140]{
		\includegraphics[width=0.23\columnwidth]{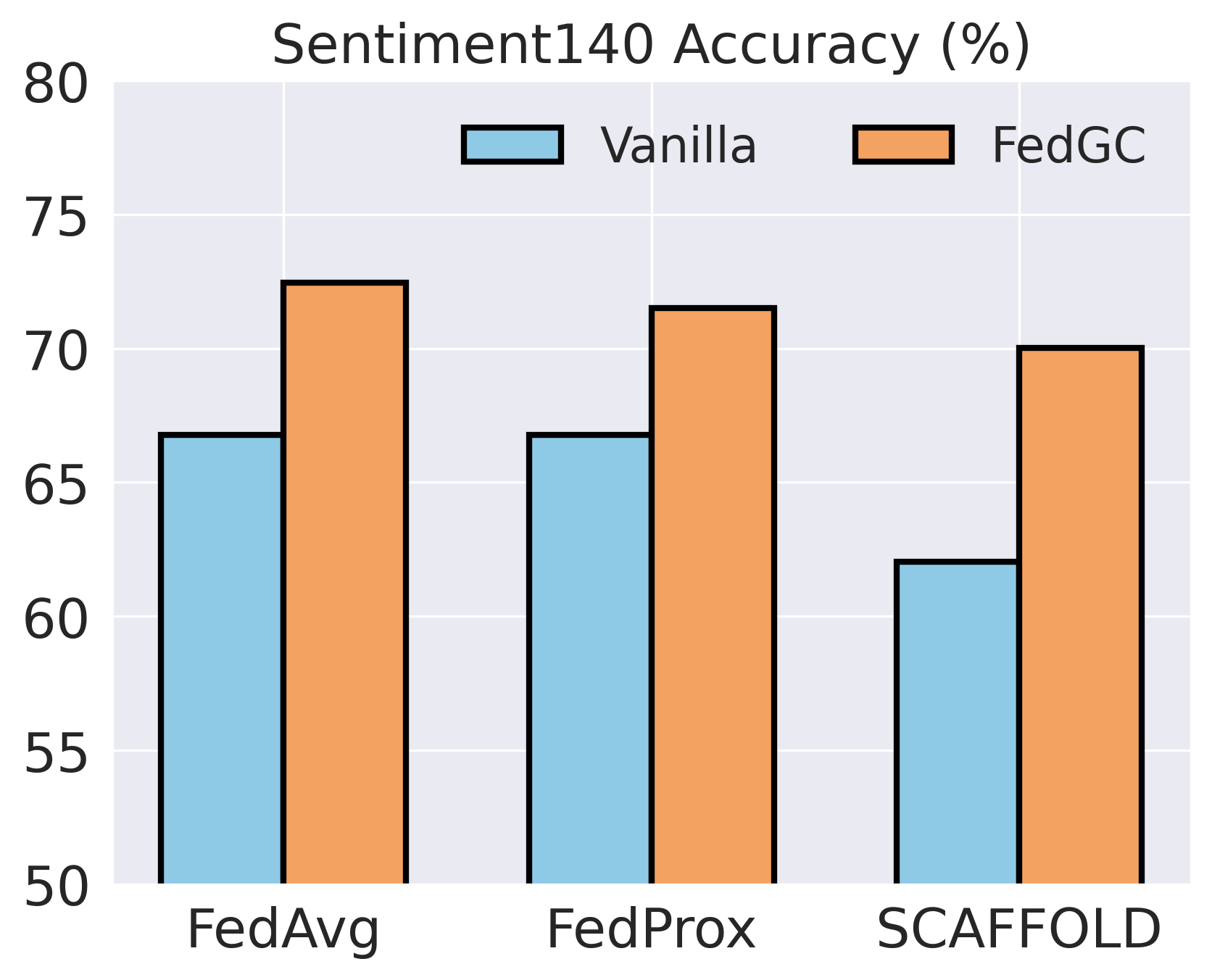}
		\label{fig:sentiment}
	}
	\subfigure[Yahoo! Answers]{
		\includegraphics[width=0.23\columnwidth]{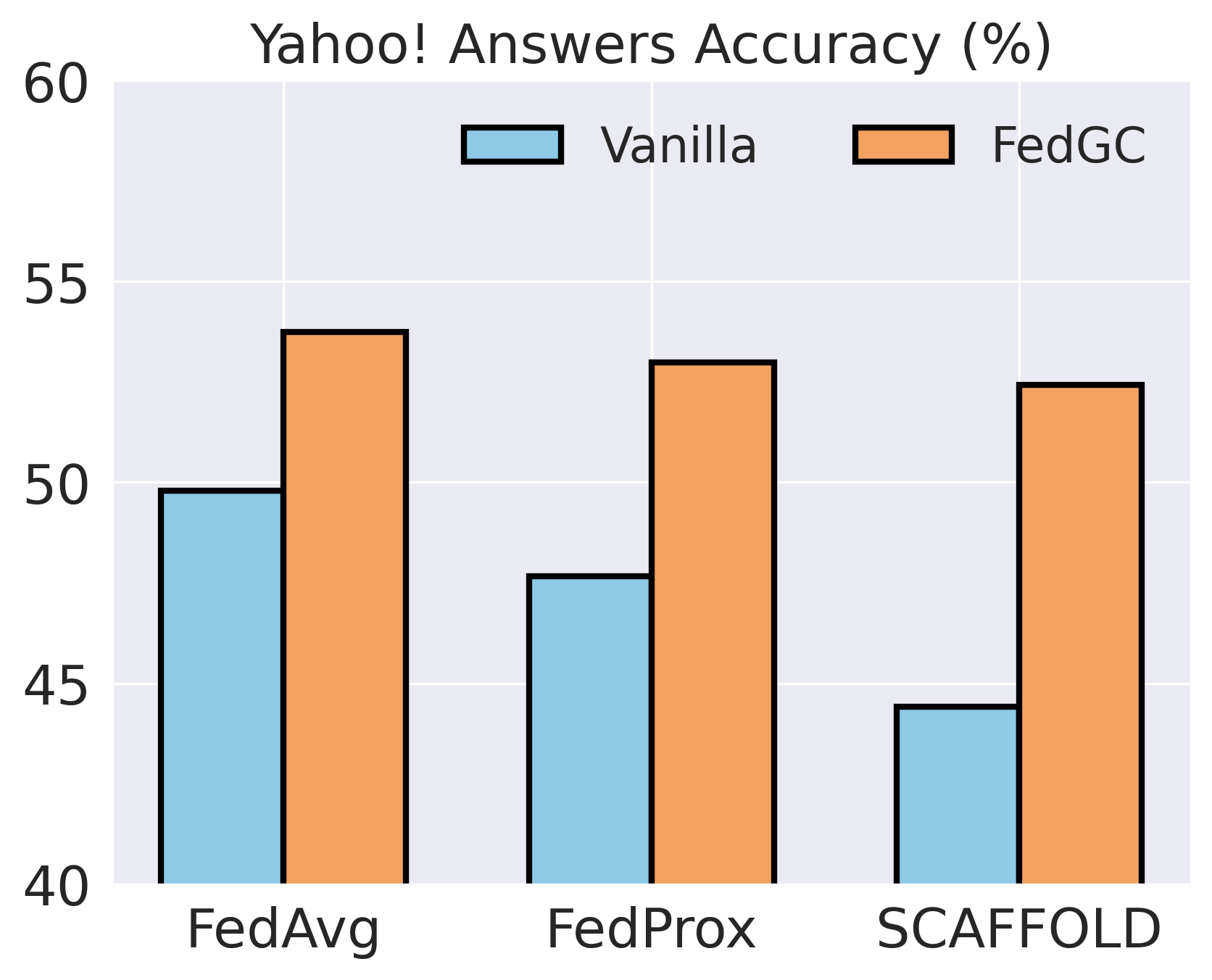}
		\label{fig:yahoo}
	}
    \vspace{-10pt}
   \caption{Performance comparisons on two text datasets. Our proposed FedGC consistently and significantly brings performance gain.}
    \vspace{-10pt}
   \label{fig:text}
\end{wrapfigure}

\textbf{FedGC is general across modalities.}
In Figure~\ref{fig:text}, we report the performance of FedGC in text modality.
We consider two datasets, Sentiment140 and Yahoo! Answers, consisting of 1000 and 100 clients, respectively.
Here, we use ChatGPT~\citep{ouyang2022training} as the generative model.
We apply equal budget allocation and single prompt (where we increase the diversity by directly instructing ChatGPT to ``be diverse'').
For real-data-guidance, we take advantage of LLM's few-shot learning ability by giving several real examples in the context~\citep{brown2020language}.
From the figure, we see that FedGC still consistently and significantly brings performance gain to all three baselines across two datasets.
This experiment verifies that our proposed FedGC framework has the potential to generalize well to diverse modalities.

\textbf{Applicability to diverse scenarios.}
We also 
(1) consider scenarios where only a partial of the clients are capable of generating data in Section~\ref{app:partial_fedgc};
(2) experiment on partial client participation scenarios in Section~\ref{app:partial_client};
(3) experiment under different heterogeneity levels in Section~\ref{app:hetero_level}.

\subsection{Towards Different Designs of FedGC} 

\textbf{Generating more data could make FedAvg prevail.} In Table~\ref{tab:num_gen}, we explore the effects of number of generated samples on FL's performance. 0 denotes vanilla FL baseline. Experiments are conducted on CIFAR-10 ($\beta=0.05$). From the table, we have an interesting finding: (1) when the number of generated samples is relatively small (0$\sim$2000), FedGC can enlarge the gap between standard FedAvg and the method (SCAFFOLD) that is specifically designed for addressing data heterogeneity; (2) however, as the number continues to grow, the situation is reversed that the basic FL method FedAvg prevails. This finding suggests that apart from carefully designing FL algorithm, it is also a promising direction to explore the greater potential from the perspective of generative data.

\begin{table}[t]
\setlength\tabcolsep{5pt}
\caption{Increasing number of generated samples makes FedAvg~\citep{fedavg} prevail.}
\label{tab:num_gen}
\begin{center}
\begin{tabular}{ccccccccccc}
\toprule
No. Gen. & 0 & 100 & 200 & 500 & 1000 & 2000 & 5000 & 10000 & 20000 & 50000 \\
\midrule
FedAvg & 61.25 & 63.67 & 66.21 & 67.13 & 66.98 & 66.28 & 71.65 & \cellcolor{blue!8}\textbf{74.50} & \cellcolor{blue!8}\textbf{76.93} & \cellcolor{blue!8}\textbf{76.39}\\
FedProx & \cellcolor{blue!8}\textbf{64.02} & 66.47 & 67.40 & 67.05 & 68.55 & 69.19 & \cellcolor{blue!8}\textbf{72.10} & \cellcolor{blue!8}\textbf{74.36} & \cellcolor{blue!8}\textbf{76.81} & \cellcolor{blue!8}\textbf{76.73}\\
SCAFFOLD & \cellcolor{blue!8}\textbf{63.98} & \cellcolor{blue!8}\textbf{69.05} & \cellcolor{blue!8}\textbf{71.33} & \cellcolor{blue!8}\textbf{71.55} & \cellcolor{blue!8}\textbf{71.33} & \cellcolor{blue!8}\textbf{70.04} & 70.34 & 73.96 & 74.88 & 73.98\\
\bottomrule
\end{tabular}
\end{center}
\end{table}

\begin{table}[t]
\begin{minipage}[t]{0.5\textwidth}
\setlength\tabcolsep{3.3pt}
\centering
\captionof{table}{Different prompt designs of FedGC applied on baselines. The design of multiple prompt formats is preferred for its effectiveness, diversity, and simplicity.}
\label{tab:prompt}
\begin{tabular}{ccc>{\columncolor{gray!15}}cc}
\toprule
Baseline   & No-GC & Single & \textbf{Multiple} & LLM \\
\midrule
FedAvg   & 27.06 & 50.53 & \textbf{54.08} & 41.32 \\
FedProx  & 29.12 & 50.48 & \textbf{53.03} & 40.82 \\
SCAFFOLD & 28.56 & 54.13 & \textbf{58.53} & 45.87 \\ 
\bottomrule
\end{tabular}
\end{minipage}
\hspace{2mm}
\begin{minipage}[t]{0.47\textwidth}
\setlength\tabcolsep{4pt}
\centering
\captionof{table}{Different generation guidance designs of FedGC applied on baselines. The mixed guidance that combines text2img and img\&text2img is the most effective strategy.}
\label{tab:guidance}
\begin{tabular}{cccc>{\columncolor{gray!15}}c}
\toprule
Baseline   & Pri. & T2I & IT2I & \textbf{Mixed} \\
\midrule
FedAvg   & 48.57 & 51.91 & 42.38 & \textbf{56.67} \\
FedProx  & 49.52 & 51.43 & 44.76 & \textbf{56.19} \\
SCAFFOLD & 54.76 & 56.67 & 49.52 & \textbf{58.57} \\ 
\bottomrule
\end{tabular}
\end{minipage}
\end{table}

\begin{table}[t!]
\begin{minipage}[t]{0.41\textwidth}
\setlength\tabcolsep{4pt}
\centering
\captionof{table}{Different budget allocation strategies of FedGC applied on baselines. Equal allocation is preferred for its effectiveness and simplicity.}
\label{tab:allocation}
\begin{tabular}{c>{\columncolor{gray!15}}ccc}
\toprule
Baseline   & \textbf{Equal} & Inverse & Water \\
\midrule
FedAvg   & \textbf{74.50} & 68.10 & 71.26 \\
FedProx  & \textbf{74.36} & 68.51 & 72.23 \\
SCAFFOLD & 73.96 & 73.94 & \textbf{74.43} \\ 
\bottomrule
\end{tabular}
\end{minipage}
\hspace{2mm}
\begin{minipage}[t]{0.55\textwidth}
\setlength\tabcolsep{4pt}
\centering
\captionof{table}{Different training strategies of FedGC applied on baselines. \textcolor{red}{Generated} data can only exhibit its efficacy when used in conjunction with real data. \textbf{Mixed} training is the most effective.}
\label{tab:training_strategy}
\begin{tabular}{ccccc>{\columncolor{gray!15}}c}
\toprule
Baseline   & Pri. & \color{red}Gen. & P2G & G2P & \textbf{Mixed} \\
\midrule
FedAvg   & 60.77 & \color{red}41.85 & 67.06 & 67.11 & \textbf{73.99} \\
FedProx  & 63.62 & \color{red}40.93 & 67.23 & 69.04 & \textbf{73.69} \\
SCAFFOLD & 65.00 & \color{red}43.45 & 66.73 & 69.50 & \textbf{75.79} \\ 
\bottomrule
\end{tabular}
\end{minipage}
\end{table}

\textbf{Equal allocation is a preferred allocation strategy for its effectiveness and simplicity.} In Table~\ref{tab:allocation}, we compare different budget allocation strategies on CIFAR-10, including equal allocation, inverse allocation, and water-filling-based allocation. Experiments show that equal allocation contributes to better performance for both FedAvg and FedProx, and comparable performance compared with water-filling-based allocation for SCAFFOLD. Considering effectiveness and simplicity, we conclude that equal allocation is a preferred allocation strategy.

\textbf{Multiple prompts lead to better performance, while LLM-based diversification might be unnecessary.} 
In Table~\ref{tab:prompt}, we explore different prompt designs on PACS dataset. PACS contains significant label-level and feature-level variations, making it an apt choice for this exploration.
We compare baseline without FedGC, FedGC with single, multiple, and LLM-based prompts (see prompt generation in Table~\ref{tab:prompt_scene}).
From the table, (1) we see that FedGC incorporated with all the prompt designs improves the performance of baselines (see improvement over the No-GC column).
(2) We see that multiple prompts consistently and significantly perform better, while LLM-based prompts perform ordinarily. 
This may result from the fact that the scene descriptions from the LLM are usually complicated, causing multifaceted patterns in one sample, thereby complicating model training.
Overall, we prefer the design of multiple prompts for its effectiveness, diversity, and simplicity.

\textbf{Mixed guidance contributes to higher performance for rare tasks.} 
In Table~\ref{tab:guidance}, we compare different generation guidance designs on a medical dataset HAM10000~\citep{tschandl2018ham10000}. 
The reason for choosing this dataset is that the diffusion model~\citep{rombach2022high} fails to correctly understand medical prompts~\citep{kazerouni2022diffusion}, which helps support our claim more convincingly. We consider three designs, including text-guided generation (T2I), our proposed data generation with guidance of text and real data (IT2I), and the \underline{mixed usage} of T2I and IT2I. These experiments convey three interesting findings: (1) even though the diffusion model fails to generate data that visually agrees with real data, the generated data still contributes to enhancing the performance of FL (see improvement from Pri. to T2I). (2) IT2I itself fails to bring performance gain, which may result from the limited diversity and incapability to generate for missing classes. (3) Mixing these two strategies contributes to consistently and significantly better performance.

\textbf{Mixed training is the most effective training strategy.} In Table~\ref{tab:training_strategy}, we compare different training strategies on CIFAR-10, including training only on the private dataset (Pri.), training only on the generative dataset (Gen.), sequential training with private dataset first (P2G), sequential training with generative dataset first (G2P), and mixed training. Experiments show that 1) generative data itself fails to ensure training, indicating that there is a gap between generative data and real private data. 2) However, when using generative data together with real private data, we see consistent performance gain compared to training on private data. This indicates that despite the incapability of fully representing real data, the generative data still contributes to improving training by increasing diversity. 3) Mixed training consistently and significantly achieves better performance.

\subsection{Towards Deeper Understanding of FedGC}

\begin{wrapfigure}{r}{4.3cm}
\vspace{-15pt}
   \centering
    \includegraphics[width=0.3\columnwidth]{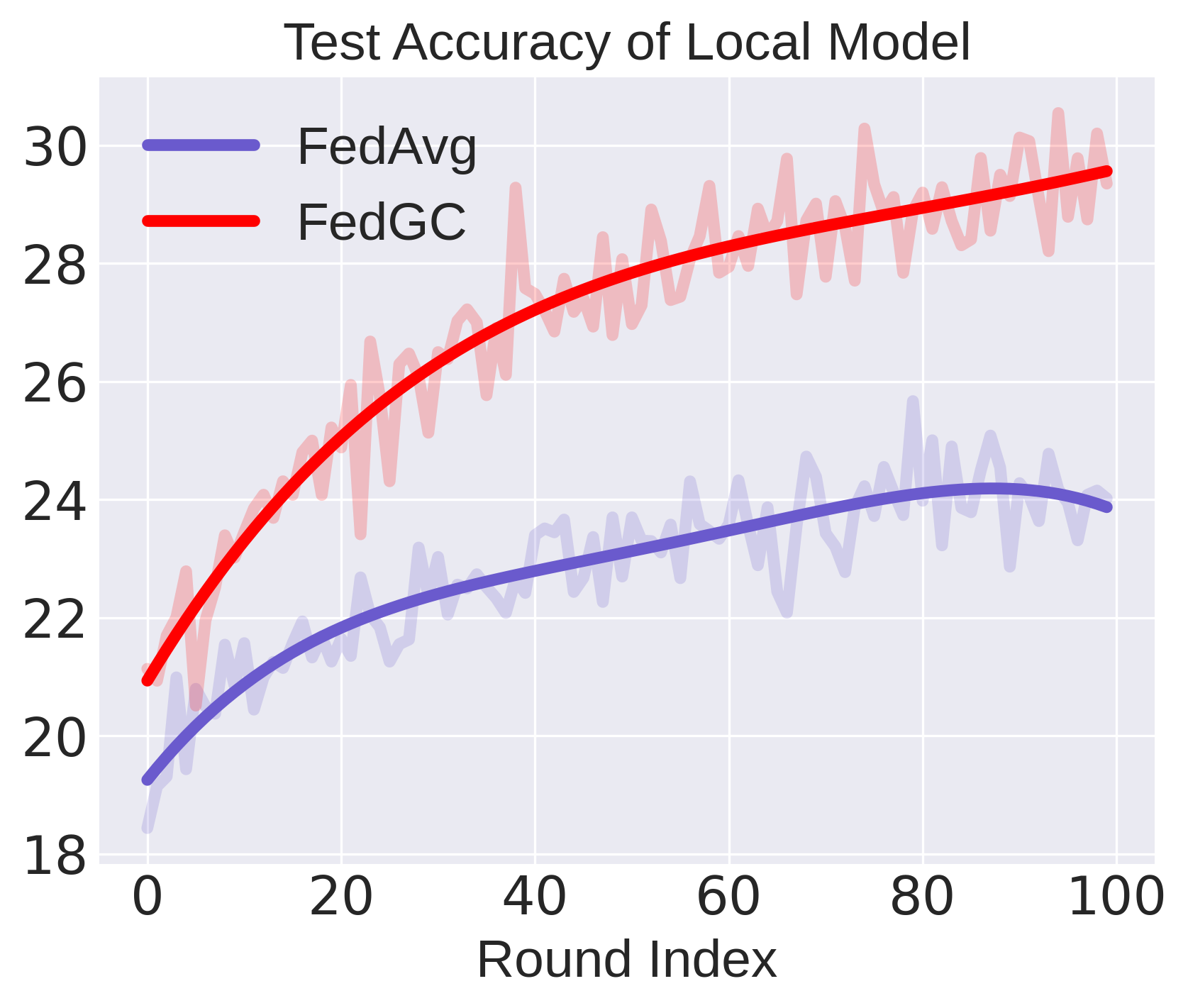}
    \vspace{-10pt}
    \caption{Local models in FedGC better preserve capability in general tasks.}
    \vspace{-15pt}
    \label{fig:vis_overfit}
\end{wrapfigure}

\textbf{Generated data is diverse, but may not be similar to real data.} In Figure~\ref{fig:vis_generated_data}, we visualize the real data and generated data on EuroSAT. 
We notice that the generated data samples do not always closely resemble real images, indicating the gap between generative data and real private data (at least visually).
Yet, their inclusion still improves the FL's performance under data heterogeneity, which may result from two perspectives.
(1) The generative data might act as a form of data augmentation, which potentially introduces variations that are not covered by the original dataset.
(2) The generative data diversify the dataset, which serves as a form of implicit regularization, preventing the model from over-fitting to the potentially biased private local data.

\textbf{FedGC alleviates over-fitting local data distribution.} In Figure~\ref{fig:vis_overfit}, we compare the averaged test accuracy of local models on the global test dataset. 
From the figure, we can see a clear accuracy gap between our FedGC and the baseline FedAvg.
(1) This indicates that our proposed FedGC can encourage each client to preserve the capability on the global general task, rather than overly fit the local specific task (local data distribution).
(2) This also helps explain why the generative data can bring performance gain even though they may fail to resemble real data.

\begin{wrapfigure}{r}{5cm}
\vspace{-15pt}
   \centering
    \includegraphics[width=0.35\columnwidth]{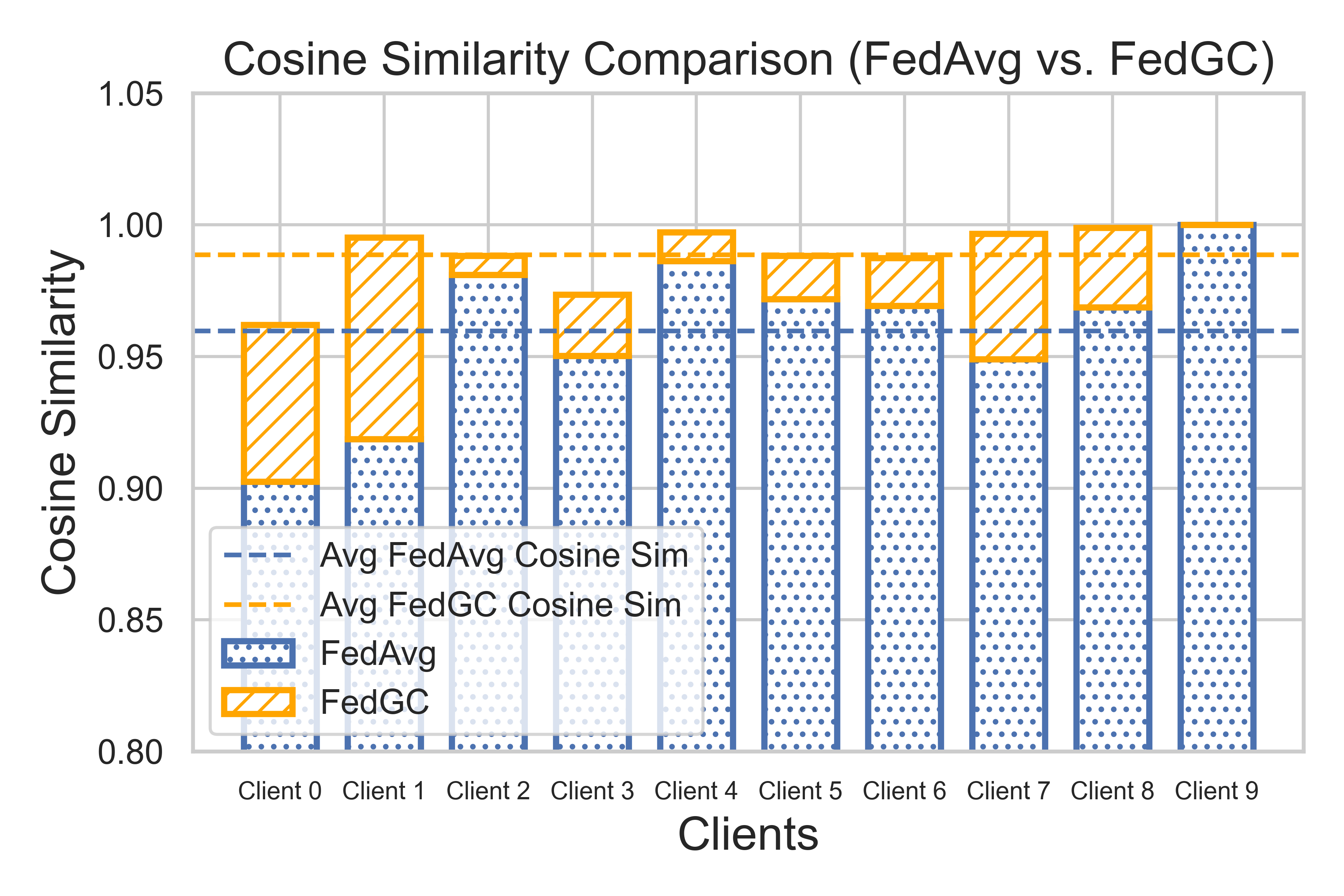}
    \vspace{-10pt}
    \caption{FedGC increases cosine similarity between local datasets (i.e., reduces data heterogeneity).}
    \vspace{-10pt}
    \label{fig:vis_feature}
\end{wrapfigure}

\textbf{FedGC reduces data heterogeneity.} 
In Figure~\ref{fig:vis_feature}, we explore the effects of FedGC on data heterogeneity from the perspective of data.
To measure the data heterogeneity, we first extract the features of data for each client using a pre-trained ResNet-18~\citep{resnet}, average the features, and compute the pair-wise cosine similarity among the averaged features of all clients.
Figure~\ref{fig:vis_feature} shows the pair-wise similarity using Client 9 as the reference.
From the figure, we consistently see that FedGC can significantly increase the similarity between datasets of two clients, verifying that FedGC can contribute to mitigating data heterogeneity.
We also report $\ell_2$ distance as metric and results on PACS in Figure~\ref{app:eurosat} and \ref{app:pacs}.

\begin{figure}[t]
	\centering
    \subfigure[Model Divergence at $\beta=0.05$]{
	   \includegraphics[width=0.31\columnwidth]{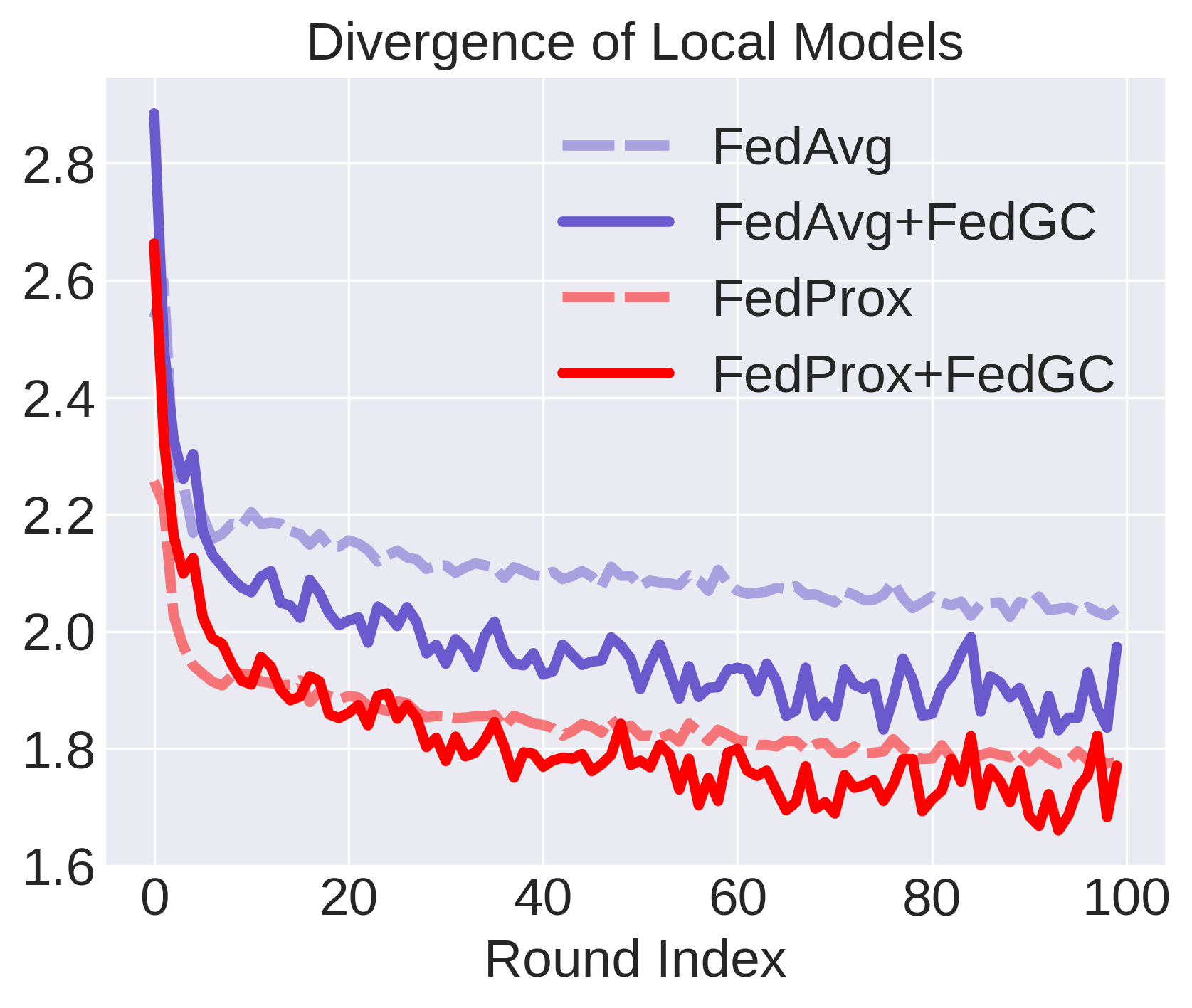}
        \label{fig:vis_div_0.05}
    }
    \subfigure[Model Divergence at $\beta=0.1$]{
		\includegraphics[width=0.31\columnwidth]{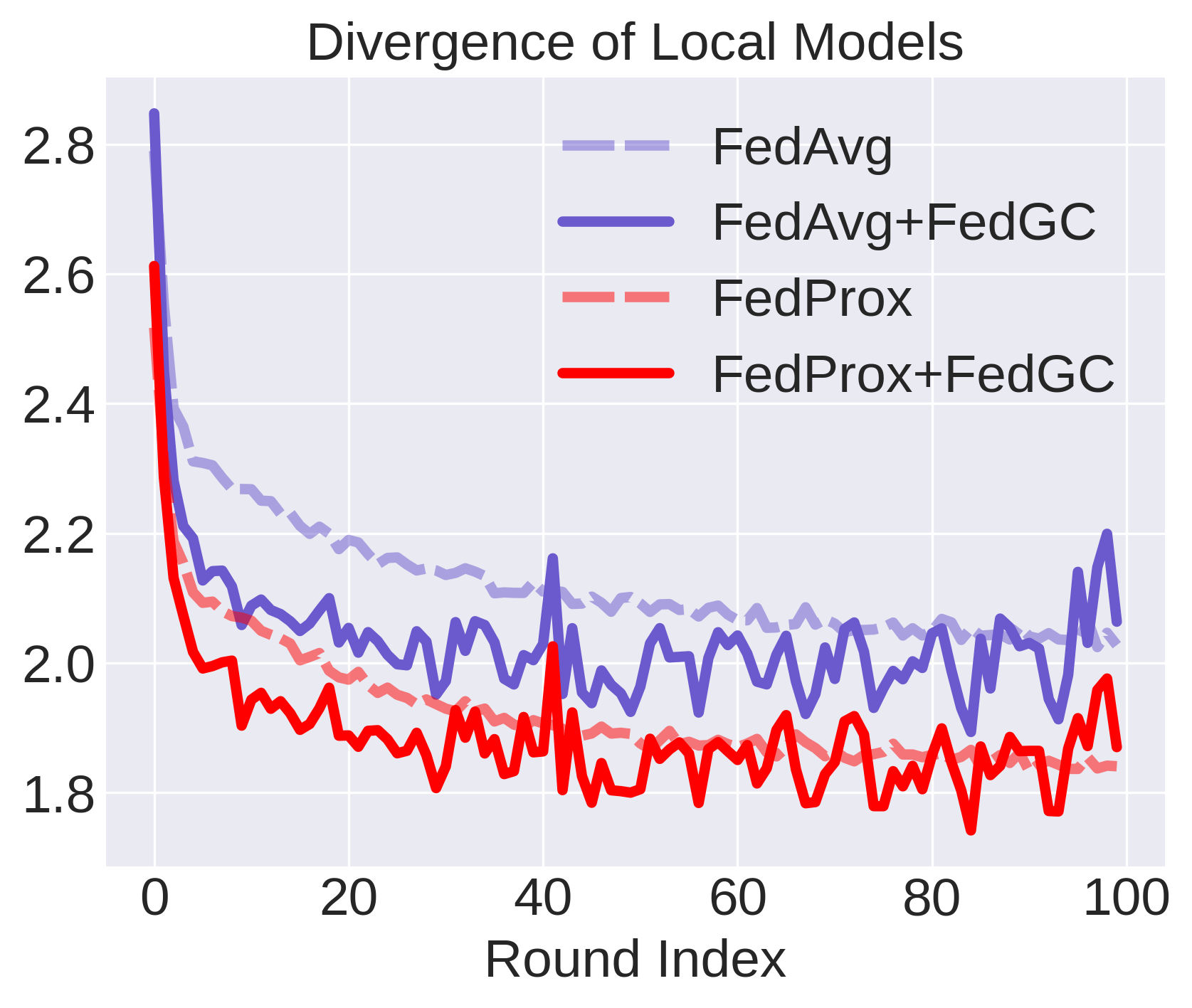}
		\label{fig:vis_div_0.1}
	}
	\subfigure[Divergence v.s. Accuracy]{
		\includegraphics[width=0.31\columnwidth]{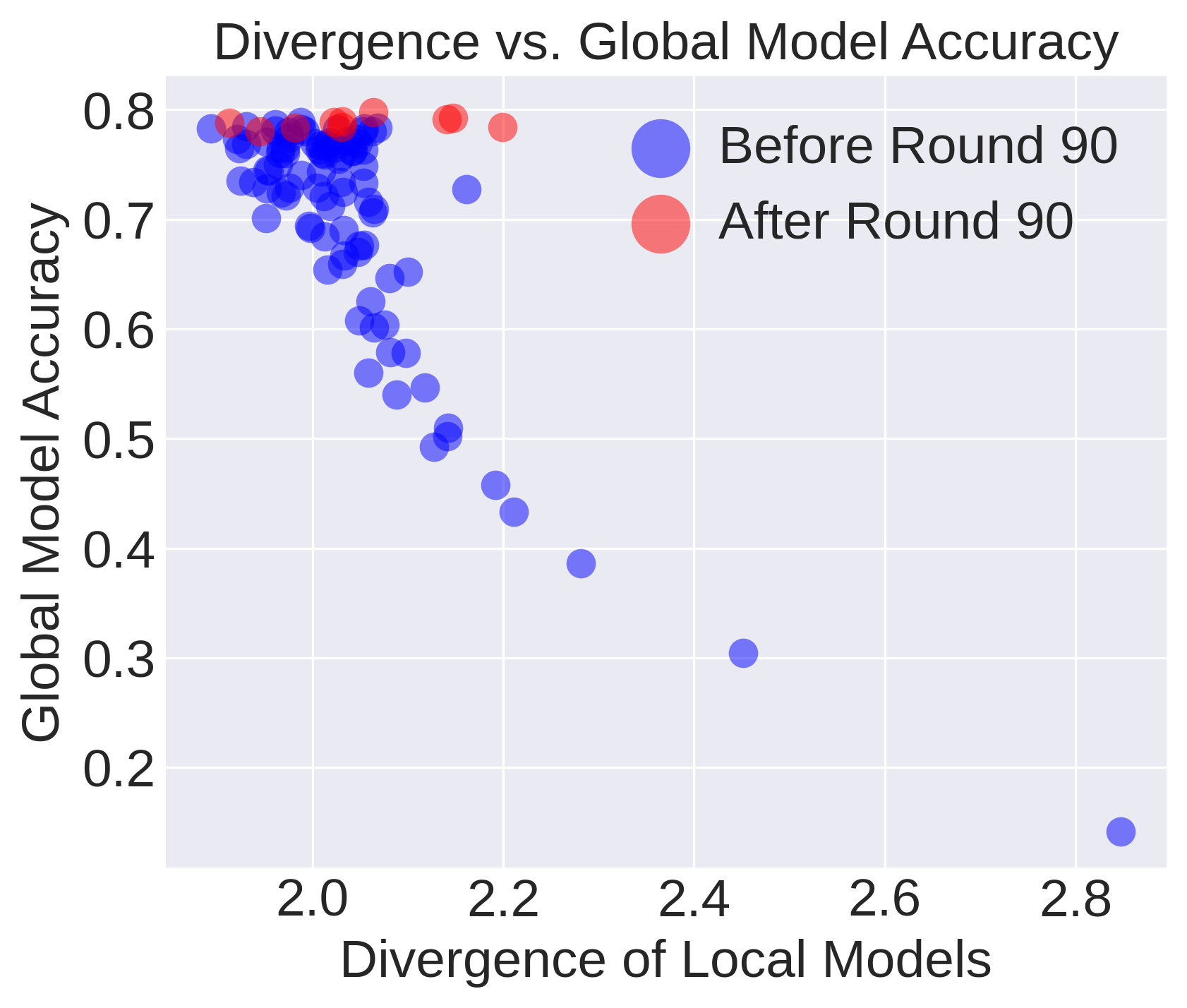}
		\label{fig:vis_acc_div}
	} \vspace{-8pt}
\caption{FedGC implicitly reduces model divergence. Notably, even with higher divergence in final rounds, we still observe higher accuracy, reflecting the potential dual nature of data diversity.}
\label{fig:vis_divergence}
\end{figure}

\textbf{FedGC implicitly reduces model divergence.} In Figure~\ref{fig:vis_divergence}, we visualize the local model divergence along with the round increases. 
Specifically, at each round, we compute the $\ell_2$ difference between each local model and the aggregated global model~\citep{fedprox} and report the averaged difference.
(1) From Figure~\ref{fig:vis_div_0.05}, we see that FedGC consistently and significantly reduces the model divergence of local models under severe heterogeneity level ($\beta=0.05$).
This result well supports the claim that FedGC is a pleasant FL framework for tackling the issue of data heterogeneity since it has been shown that data heterogeneity leads to larger model divergence and thus mediocre performance empirically~\citep{fedprox} and theoretically~\citep{wangsurvey,li_survey}.

(2) From Figure~\ref{fig:vis_div_0.1}, though the divergence of FedGC increases at the final rounds, we still observe improved accuracy during these rounds.
This observation is interesting as it seems to contradict the current viewpoint. 
Based on this, we hypothesize that data diversity has two sides: 
(1) it can reduce data heterogeneity, thus reducing model divergence;
(2) but it can also increase model divergence as the data is more diverse.
Nevertheless, data diversity still contributes to enhanced model performance despite the increased divergence as suggested in Figure~\ref{fig:vis_acc_div} (see the red scatters).
This interesting finding calls for more theoretical future works to model data diversity rather than only model divergence in the FL theory~\citep{scaffold,fednova}.

\section{Discussions and Future Directions}

As an initial exploration of tackling data heterogeneity in FL using generative content, we mainly focus on designing under two standards: simplicity and effectiveness.
However, there are also diverse interesting future directions that are worth exploring.

\textbf{Filtering.}
Previously, we find that the positive effects of data diversity on data heterogeneity outweigh the negative effects of some low-quality generated samples and the gap between generated and real data.
Still, it could benefit if we can propose to appropriately filter out low-quality samples, such as applying the KNN method~\citep{fix1989discriminatory} using real data as reference.
We propose an initial attempt using the global model as a discriminator to filter data in Section~\ref{app:filter}.

\textbf{Personalized generation.}
It would also be advantageous to narrow the gap between generated and real data.
Beyond our proposed real-data-guided generation, delving into personalized data generation to more closely resemble real data is worth investigating.
For example, we can fine-tune generative models on real data using parameter-efficient fine-tuning techniques~\citep{hu2021lora}.

\textbf{Theory.}
As shown in Figure~\ref{fig:vis_div_0.1} and \ref{fig:vis_acc_div}, model divergence~\citep{li2019convergence,fedprox,wangsurvey} may not fully represent the property of distributed data, calling for more future theory works.

\section{Limitations and Conclusion}

\textit{Limitations and Future Works.} Despite putting much effort into diversifying the experimental settings, there are still cases not covered. 
For example, we only explore one diffusion model and LLM respectively. 
There could be future works to explore the effects of different generative models.

\textit{Conclusion.} 
This paper focuses on the notorious issue of data heterogeneity in FL.
We propose a new, simple yet effective FL framework termed FedGC, which leverages generative data to promote FL under heterogeneous private data.
We summarize four crucial aspects that are worth exploring and propose three solutions for each aspect.
Extensive experiments show that our FedGC framework can consistently and significantly improve the performance of diverse FL baselines under data heterogeneity.
Moreover, we provide a systematic empirical analysis based on FedGC and provide new insights throughout the experimental section.
Our research serves as an initial exploration of boosting federated learning on private data in the era of generative content.

\section*{Acknowledgement}

This research is supported by the National Key R\&D Program of China under Grant 2021ZD0112801, NSFC under Grant 62171276 and the Science and Technology Commission of Shanghai Municipal under Grant 21511100900 and 22DZ2229005.



\bibliography{iclr2024_conference}
\bibliographystyle{iclr2024_conference}

\newpage

\appendix
\section{Appendix}

\subsection{More Illustration of FedGC}

For the prompts conditioned on the latent diffusion model, we show the LLM-based prompts for generating images in Table~\ref{tab:prompt_scene}.
In detail, we instruct ChatGPT through System Prompt and User Prompt, to help us create text samples containing the corresponding class name for image generation. Utilizing ChatGPT's rich imagination of scenarios and the diversity of text styles, we can achieve a diversity of prompts. Therefore, it helps Stable-diffusion to generate diverse and more realistic pictures.

\begin{figure}[t]
\begin{center}
\includegraphics[width=0.95\columnwidth]{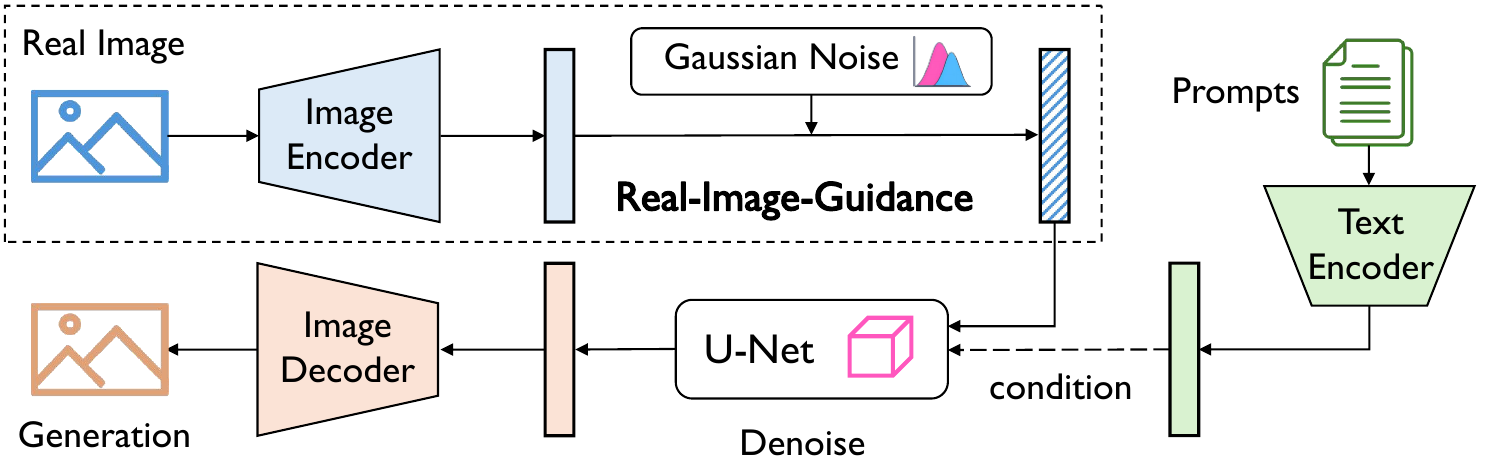}
\end{center}
\caption{Real-data-guidance for image generation based on diffusion model. The real-data-guidance method involves 4 steps: (1) initializing latent features with real-image data, (2) adding controlled noise, (3) denoising with text features, and (4) generating new images using the decoder.}
\label{fig:real-data-guide-img}
\end{figure}

For generation guidance beyond prompts, we show the real-data guidance for image generation using diffusion models in Figure~\ref{fig:real-data-guide-img}. First of all, the latent features are meticulously initialized using actual real-image data. 
Subsequently, controlled noise is introduced into the latent representations, which serves to perturb and diversify the features while maintaining the underlying structure. Following this, with conditioned prompts, we denoise this combined feature using U-Net~\citep{u-net}. Finally, passing through the image decoder, we obtain generated images.

We show the real-data-guidance for text generation using ChatGPT in Table~\ref{tab:real-data-guide-text}.
Compared to prompts containing class num, here we instruct ChatGPT to imitate the theme and content of the corresponding text and directly expand the amount of text data.
In our illustrative examples shown in Table~\ref{tab:real-data-guide-text}, we simulate real-world data scenarios by incorporating four actual instances and generating an additional set of four synthetic instances. In this experimental setup, we task ChatGPT with the generation of data that exhibits diverse patterns akin to those found in authentic real data. Furthermore, we guide ChatGPT to produce two distinct samples for each distinct label category, fostering a balanced and representative dataset.

\begin{table}[t]
\caption{Obtaining LLM-based prompts for generating images using diffusion models. Instructions for generating scene descriptions (i.e., prompts for diffusion models) given a class name using ChatGPT. Here, we provide an example on the dog category of PACS dataset.}
\label{tab:prompt_scene}
\begin{response}
\textbf{System Prompt:}\\
You are an AI assistant that helps people find information.\\
\\
\textbf{User Prompt:}\\
Please help me come up with scene descriptions that contain a dog while not containing an elephant, giraffe, guitar, horse, house, person.\\
\\
For example:\\
\\
$\left[ \text{``A dog is running on the grass'', ``A dog is sleeping on the floor''} \right]$\\
\\
Please generate 10 samples in the format of a list.\\
Remember: each description should be within 10 words. 
\end{response}
\end{table}

\begin{table}[t]
\caption{Real-data-guidance for text generation using ChatGPT. Real data is modeled in the examples, where we provide four real examples and generate four new examples. We instruct the ChatGPT to generate diverse data that has a similar pattern to real data. We also instruct the ChatGPT to generate two samples for each label.}
\label{tab:real-data-guide-text}
\begin{response}
\textbf{System Prompt:}\\
Assistant is an intelligent chatbot designed to help users generate similar data. Users will provide a few real samples and the Assistant will generate data that follows the pattern of real samples. This is a binary dataset on sentiment analysis, where 0 denotes negative and 1 denotes positive.\\
\\
Instructions:\\
1. Generate two samples with label 0 and two samples with label 1, try to make the content diverse\\
2. Should have a similar pattern of users' data.\\
\\
\textbf{User Prompt:}\\
**Data: \{example\_input\_1\}, Label: \{example\_label\_1\}**\\
**Data: \{example\_input\_2\}, Label: \{example\_label\_2\}**\\
**Data: \{example\_input\_3\}, Label: \{example\_label\_3\}**\\
**Data: \{example\_input\_4\}, Label: \{example\_label\_4\}**\\
\\
Generate two samples with label 0 and two samples with label 1.\\
In the format of Data: \{\}, Label: \{\}. Each sample should start with ** and end with **.
\end{response}
\end{table}

\subsection{Implementation Details}
\label{app:imp}

We list the number of clients for each dataset in Table~\ref{tab:client_number}.

\begin{table}[t]
\setlength\tabcolsep{5pt}
\caption{Number of clients for each dataset.}
\label{tab:client_number}
\begin{center}
\begin{tabular}{ccccccccccc}
\toprule
Dataset & CIFAR-10 & EuroSAT & PACS & VLCS & HAM10000 & Sentiment & Yahoo!\\
\midrule
Client Number & 10 & 10 & 20 & 20 & 10 & 1000 & 100\\
\bottomrule
\end{tabular}
\end{center}
\end{table}

\subsection{Membership Inference Attack}
\label{app:mia}
To measure the privacy preservation of FedAvg and FedGC, we carry out a simple membership inference attack based on loss evaluation, as \citep{sablayrolles2019white} has shown that it is reasonable to use the loss of the model to infer membership. We consider a scene where an attacker who has a tiny amount of training data can get the global model and wants to figure out whether a similar datum (i.e. also a photo of an airplane) has been used to train the model or not. During the attack, the attacker feeds its few data to the global model and trains a binary classifier based on the loss of each training-used and not-training-used datum. 

We conduct our experiment on CIFAR-10 dataset. In the training process, we set the client number to 10 and the Dirichlet distribution parameter to $\beta=0.1$. We also discard data augmentations (i.e. flipping and cropping) for more clear comparisons. In the main body, we compare both task accuracy and attack accuracy, as shown in Figure~\ref{fig:privacy_acc}. 

We also compare the attack accuracy at the point when FedAvg and FedGC achieve similar task accuracy in Table~\ref{tab:mia}.
From the table, we see a much more significant reduction in privacy leakage (i.e., much lower attack accuracy).
This is reasonable as FedGC can accelerate the convergence speed, which means FedGC requires fewer steps of optimization on the sensitive private data to achieve the same.

\begin{table}[t]
\caption{Membership inference attack accuracy comparisons when FedAvg and FedGC achieve similar task accuracy. We consider two scenarios where the total number of clients' real samples is 50k and 10k, respectively. We also explore the effects of using different number of generated samples. FedGC can reduce privacy leakage to a very low level (since random guess is 50\%) while maintaining task accuracy at the same time. }
\label{tab:mia}
\begin{center}
\begin{tabular}{cc|cc|cc}
\toprule
\multicolumn{2}{c|}{Number of Real Samples} & \multicolumn{2}{c|}{50k} & \multicolumn{2}{c}{10k} \\
\multicolumn{2}{c|}{Accuracy} & Task  &\multicolumn{1}{c|}{Attack}& Task  & \multicolumn{1}{c}{Attack}  \\
\midrule
\multirow{6}{*}{No. of Generated Samples} & 0 & 59.71 & 60.55 & 35.48 & 77.55 \\

& 10k & 61.65 & 52.05 &  35.97 & 52.80   \\

& 20k & 62.49 & 51.20 & 39.18 & 52.85   \\

& 30k & 61.82 & 51.95 & 39.40 & 52.50  \\

& 40k & 60.38 & 51.20 & 37.17 & 52.75\\

& 50k & 62.49 & 51.60 & 38.68 & 52.35\\

\bottomrule
\end{tabular}
\end{center}
\end{table}

\begin{figure}[t]
	\centering
	\subfigure[EuroSAT: real data samples]{
		\includegraphics[width=1.0\columnwidth]{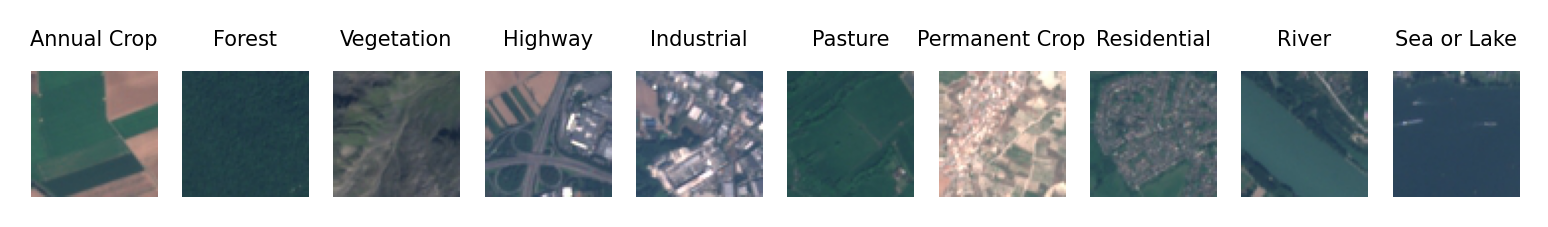}
		\label{fig:vis_eurosat_real}
	}
	\subfigure[EuroSAT: generated similar samples]{
		\includegraphics[width=1.0\columnwidth]{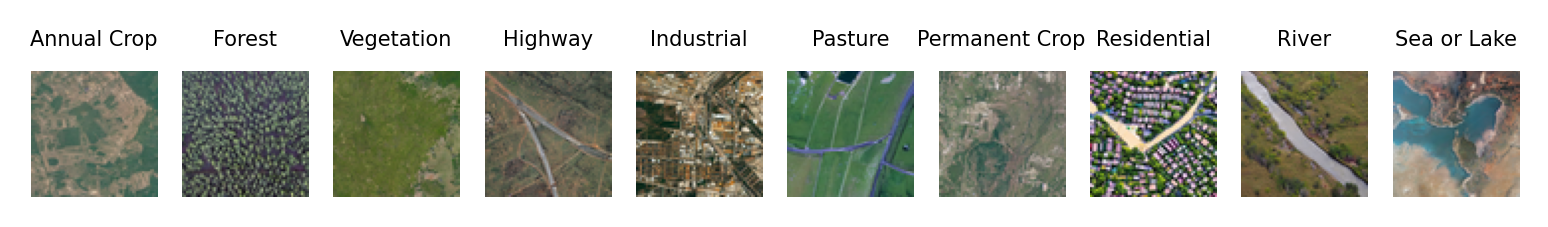}
		\label{fig:vis_eurosat_good}
	}
    \subfigure[EuroSAT: generated dissimilar samples]{
		\includegraphics[width=1.0\columnwidth]{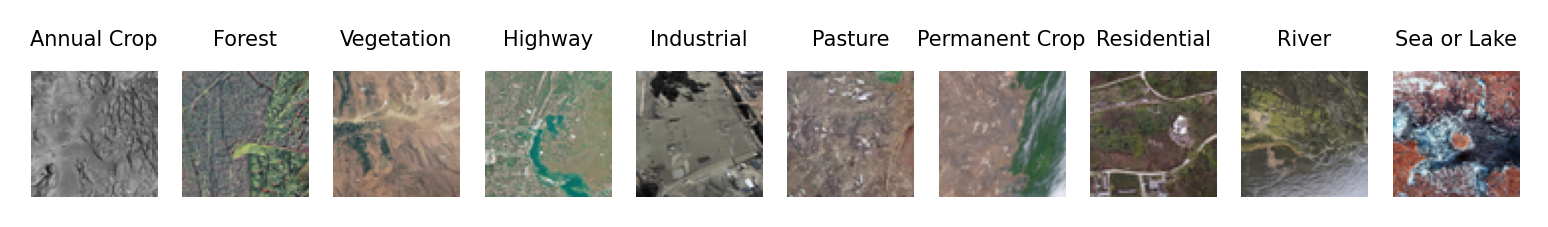}
		\label{fig:vis_eurosat_bad}
	}
	\caption{Visualization of real and generated data. (a) Visualization of real data samples from the EuroSAT dataset. (b) Visualization of generated data samples that are more aligned with the corresponding semantic or real data. (c) Visualization of generated data samples that are not aligned with the corresponding semantic or real data.}
	\label{fig:vis_generated_data}
\end{figure}

\subsection{Visualization of Real and Generated Data}
\label{app:vis_data}

We visualize the real data and generated data on EuroSAT~\citep{helber2019eurosat} in Figure~\ref{fig:vis_generated_data}. For the uncommon and detailed satellite images in EuroSAT~\citep{helber2019eurosat}, the quality of the data generated by the diffusion models varies. From the naked eye, the data generated by some diffusion can capture the semantic information brought by the label very well. For example, the generated images with the label "River" as guidance do contain rivers, but hard to achieve a similar satellite style to actual images.
Although the gap between generated and actual data definitely exists, generated data obviously improves specific task performance, which is demonstrated by our extensive experiments. 

\subsection{FedGC Mitigates Data Heterogeneity}
\label{app:feature_cos}
\begin{figure}[t]
    \centering
        \subfigure[cosine: FedAvg]{
		\includegraphics[width=0.23\columnwidth]{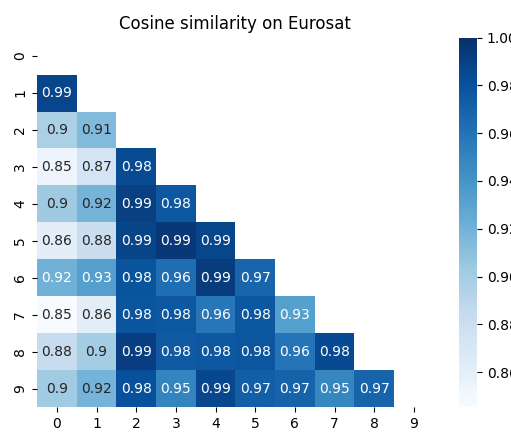}
		\label{fig:eurosat_cs}
	}
    \subfigure[cosine: FedGC]{
		\includegraphics[width=0.23\columnwidth]{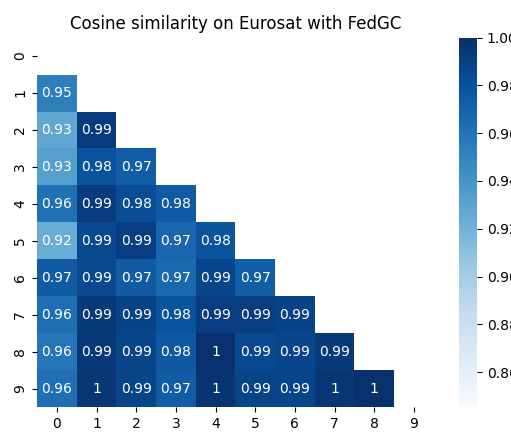}
		\label{fig:eurosat_cs_fedgc}
	}
        \subfigure[$\ell_2$: FedAvg]{
		\includegraphics[width=0.23\columnwidth]{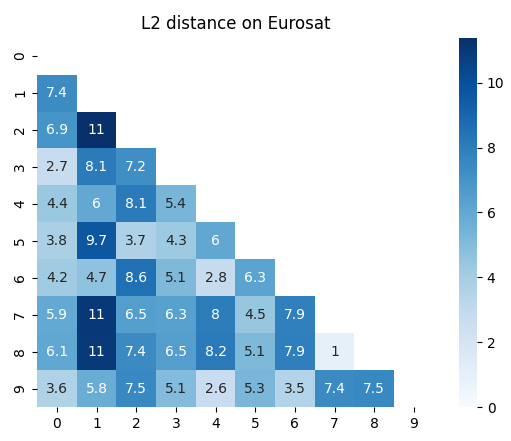}
		\label{fig:eurosat_l2}
	}
    \subfigure[$\ell_2$: FedGC]{
		\includegraphics[width=0.23\columnwidth]{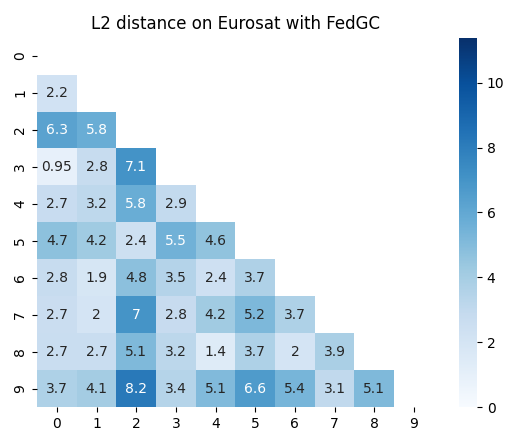}
		\label{fig:eurosat_l2_fedgc}
	}
 \caption{Feature cosine similarity and $\ell_2$ distance heatmap among 10 clients on EuroSAT. We calculate the two metrics on average data features among clients using the pre-trained ResNet-18~\citep{resnet}. FedGC enhances the feature similarity and closes their distance, which effectively mitigates the feature-level heterogeneity on EuroSAT.}
 \label{app:eurosat}
\end{figure}

\begin{figure}[t]
    \centering
        \subfigure[cosine: FedAvg]{
		\includegraphics[width=0.23\columnwidth]{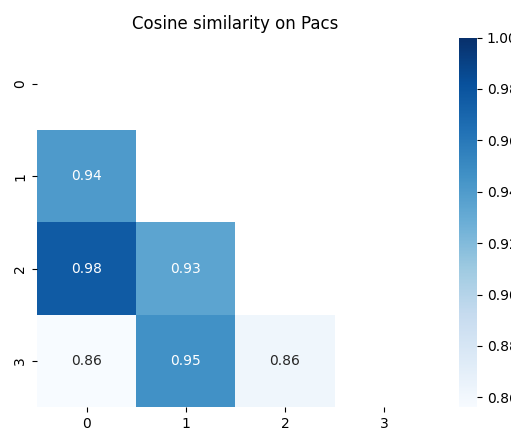}
		\label{fig:pacs_cs}
	}
    \subfigure[cosine: FedGC]{
		\includegraphics[width=0.23\columnwidth]{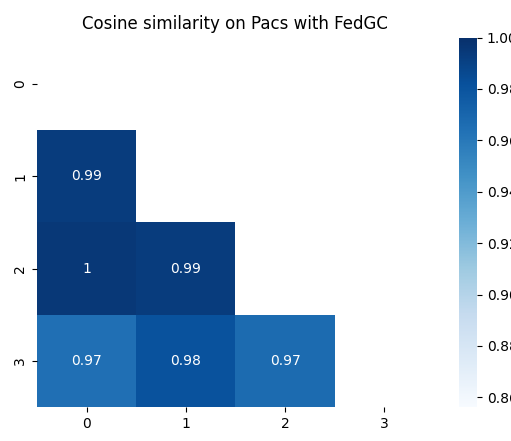}
		\label{fig:pacs_cs_fedgc}
	}
        \subfigure[$\ell_2$: FedAvg]{
		\includegraphics[width=0.23\columnwidth]{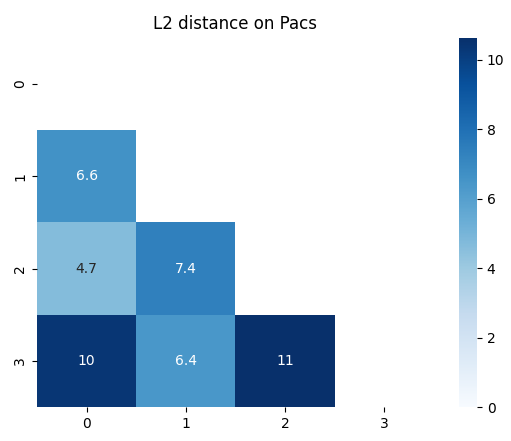}
		\label{fig:pacs_l2}
	}
    \subfigure[$\ell_2$: FedGC]{
		\includegraphics[width=0.23\columnwidth]{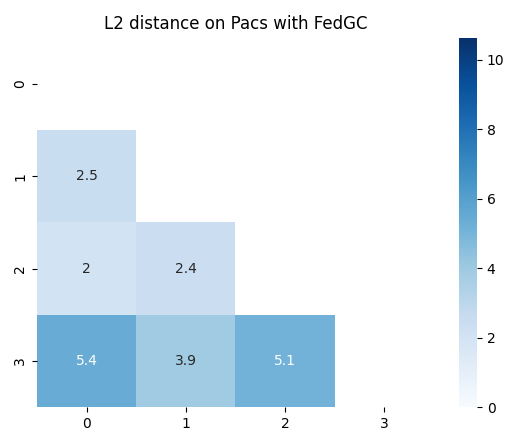}
		\label{fig:pacs_l2_fedgc}
	}
 \caption{Feature cosine similarity and $\ell_2$ distance heatmap among 4 clients on PACS. We calculate the two metrics on average data features among clients using the pre-trained ResNet-18. FedGC enhances the feature similarity and closes their distance, which effectively mitigates the feature-level heterogeneity on PACS.}
 \label{app:pacs}
\end{figure}

We visualize the cosine similarity and $\ell_2$ distance of features on EuroSAT and PACS in Figure~\ref{app:eurosat} and Figure~\ref{app:pacs} respectively.
We measure the discrepancy among local data in clients on the feature level, using 2 metrics: cosine similarity and $\ell_2$ distance. 
To be specific, we calculate the average features with pre-trained ResNet-18~\citep{resnet} on each client in turn, and then measure the indicators between all pairs of clients.

Results in the figures manifest that after applying FedGC, the cosine similarity and $\ell_2$ distance among client pairs separately increase and decrease. In other words, local data possessed by clients are more homogeneous than before. FedGC efficiently mitigates data heterogeneity by generating corresponding data on the client side. From the feature respective, we show the latent reason for significant performance improvement brought by FedGC.

\subsection{FedGC with Partial Clients Capable of Generation}
\label{app:partial_fedgc}

Our proposed FedGC framework is also applicable in cases where not every client has the capability to generate data. Here, we experiment on CIFAR-10 under two different heterogeneity levels. In Table~\ref{tab:partial_fedgc}, we compare vanilla baseline with no generative data, FedGC where all clients can generate data, and FedGC where only half of the clients can generate data.

From the table, we see that
(1) our proposed FedGC can consistently and significantly achieve the best performance despite the amount of generation-capable clients.
(2) Surprisingly, we find that under low heterogeneity level, when applied to SCAFFOLD~\citep{scaffold}, FedGC with few generation-capable clients even performs better.
This interesting finding demonstrates that our framework may be further improved by more fine-grained designs regarding who is responsible for data generation and the volume of data to be generated.

\begin{table}[t]
\caption{Experiments of a scene in which partial clients are capable of generation. 1k/50\% indicates only half of the clients are capable of generation. However, FedGC still significantly outperforms the baseline with no generative data.}
\label{tab:partial_fedgc}
\begin{center}
\begin{tabular}{c|ccc|ccc}
\toprule
 H-Level & \multicolumn{3}{c|}{High} & \multicolumn{3}{c}{Low} \\
Generation & No  &\multicolumn{1}{c}{ 1k/100\%}&\multicolumn{1}{c|}{1k/50\%} & No  & \multicolumn{1}{c}{1k/100\%} & \multicolumn{1}{c}{1k/50\%} \\
\midrule
FedAvg &  60.77 & 73.99 &  71.53 & 71.57 & 79.73 & 77.45   \\

FedProx &  63.62 & 73.69 & 72.65 & 75.76 & 79.25 & 79.23  \\

SCAFFOLD &  65.00 & 75.75 & 73.28 & 78.74 & 80.29 & 81.27  \\

\bottomrule
\end{tabular}
\end{center}
\end{table}

\subsection{FedGC under Different Heterogeneity Levels}
\label{app:hetero_level}
\begin{figure}[t]
    \centering
        \subfigure[FedAVg]{
		\includegraphics[width=0.31\columnwidth]{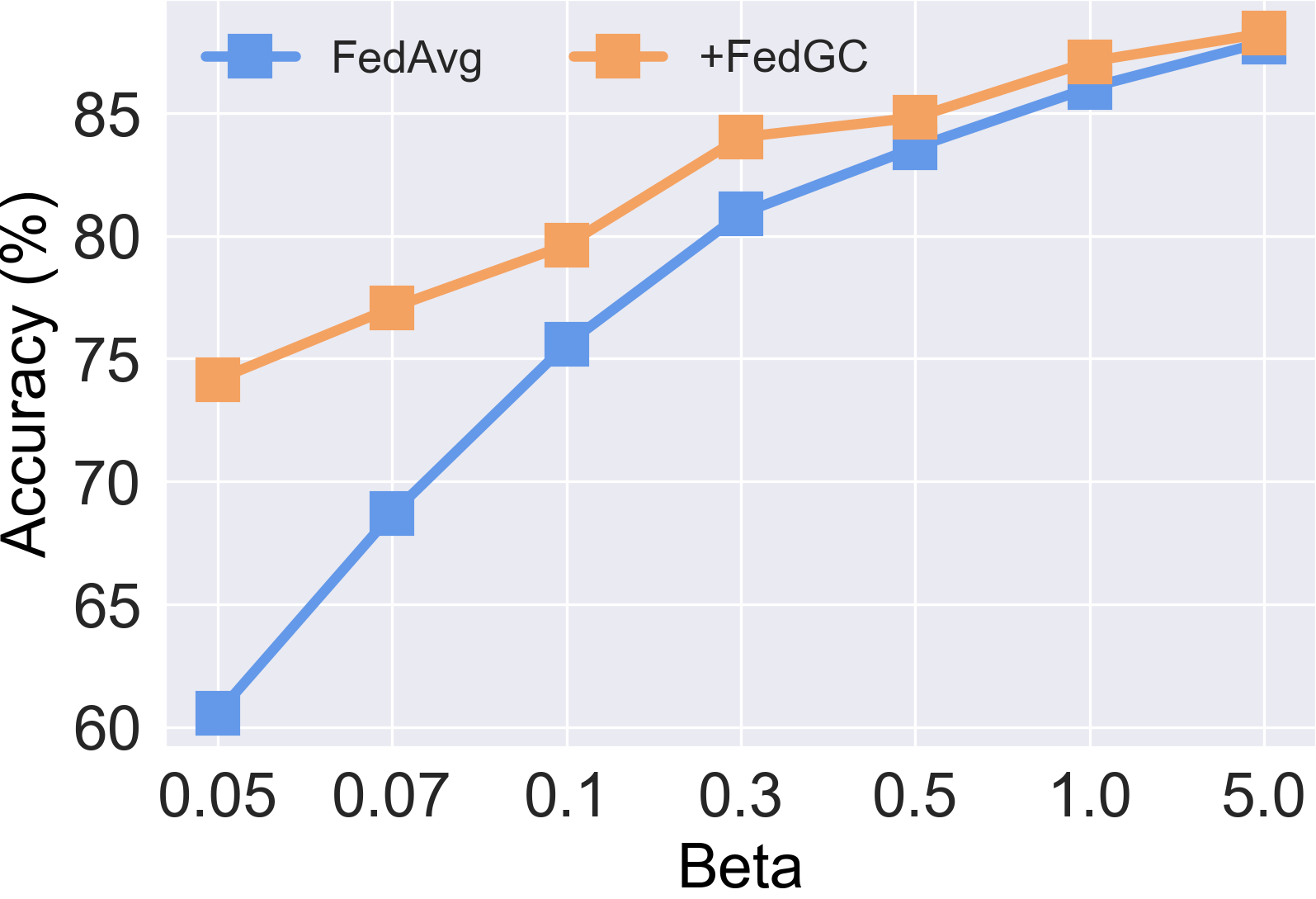}
		\label{fig:hetero-level_fedavg}
	}
    \subfigure[FedProx]{
		\includegraphics[width=0.31\columnwidth]{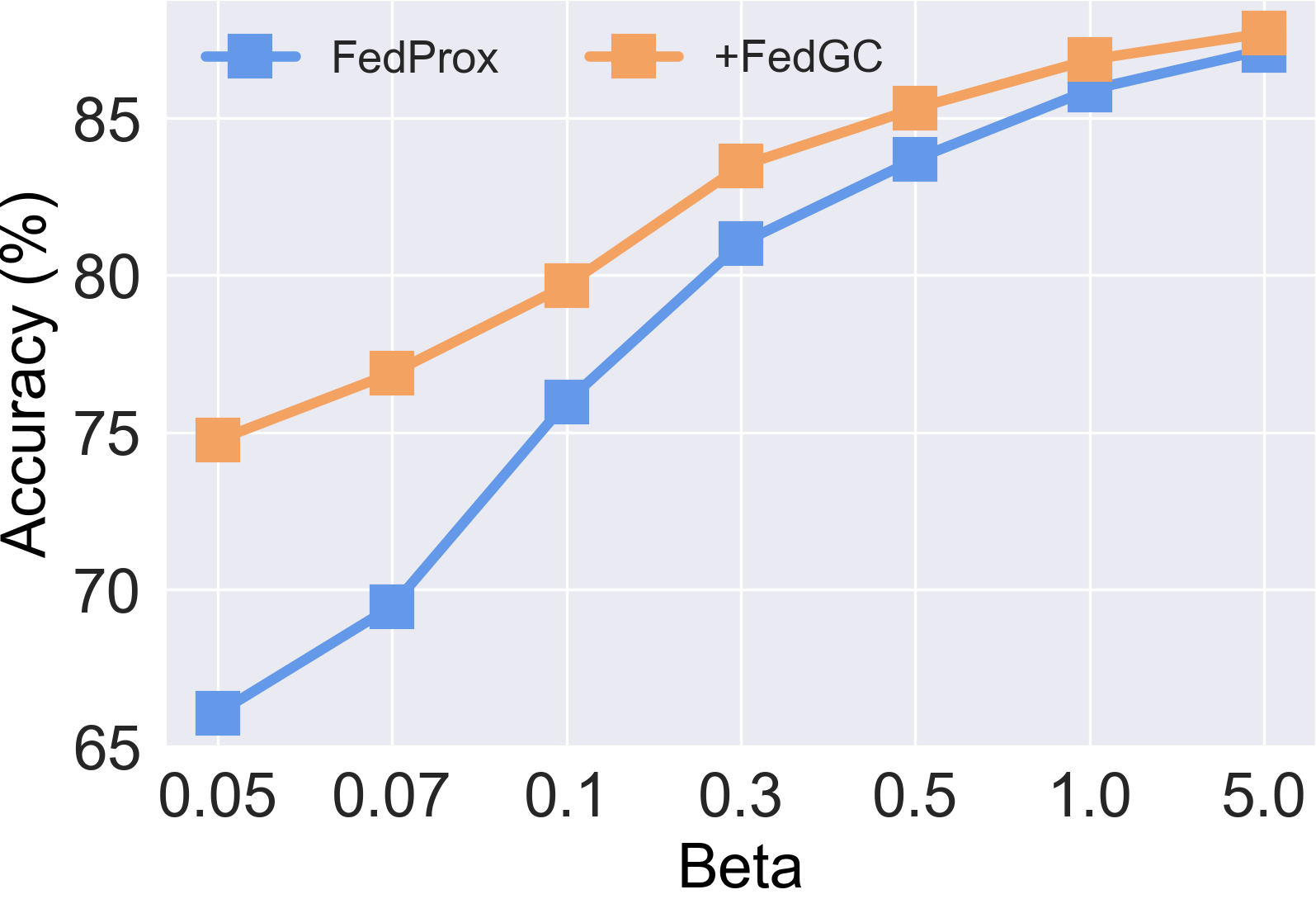}
		\label{fig:hetero-level_fedprox}
	}
    \subfigure[SCAFFOLD]{
		\includegraphics[width=0.31\columnwidth]{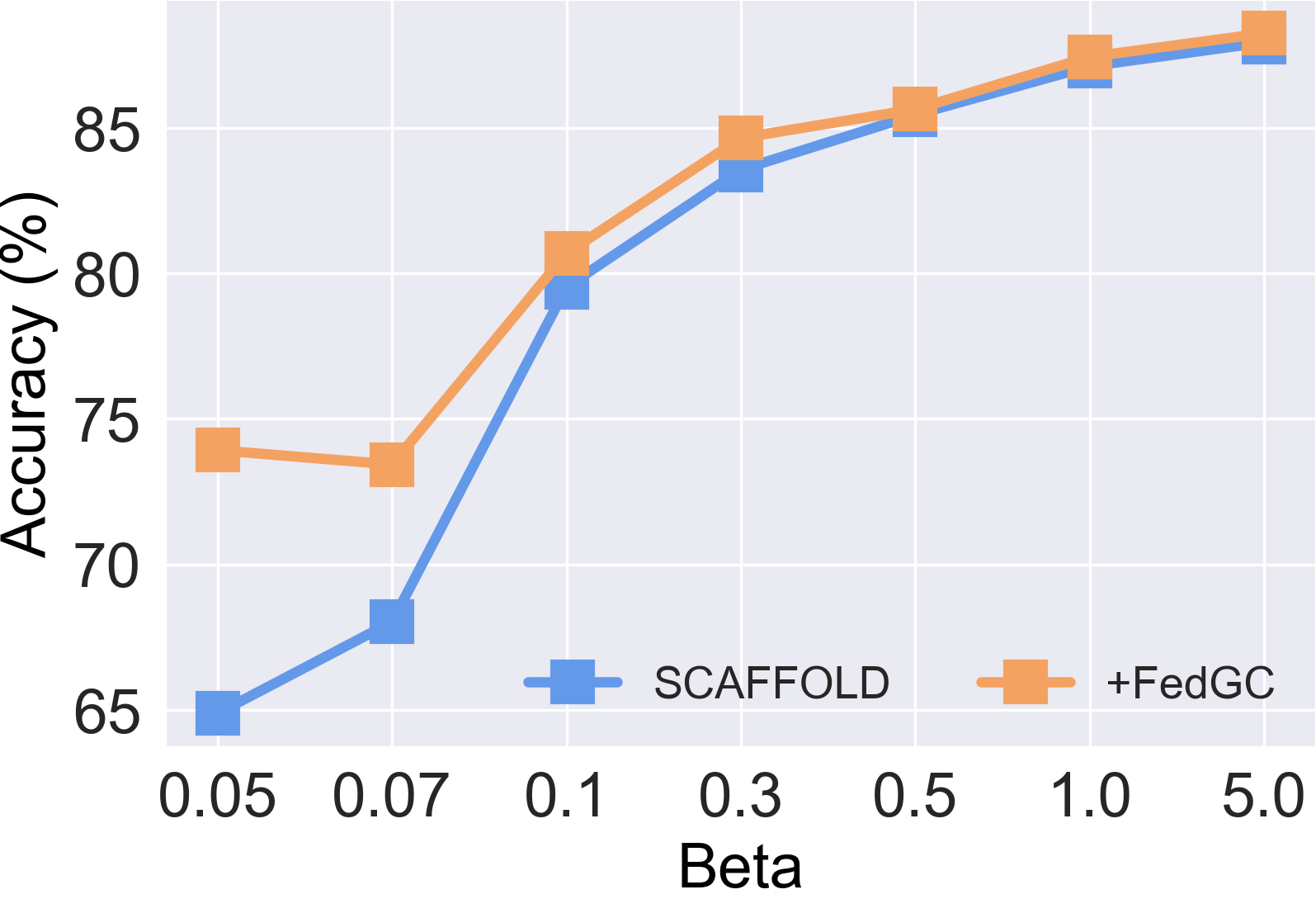}
		\label{fig:hetero-level_scaffold}
	}
 \caption{Performance comparisons between vanilla baseline and baseline in FedGC framework under different heterogeneity levels on CIFAR-10. Beta ($\beta$) is the hyper-parameter in Dirichlet distribution. As the heterogeneity level increases (Beta decreases), the improvement brought by FedGC becomes more significant. This indicates that FedGC can effectively alleviate the issue of data heterogeneity. }
 \label{fig:hetero-level}
\end{figure}

Here, we conduct experiments of three baselines including FedAvg, FedProx, and SCAFFOLD, with different heterogeneity levels on CIFAR-10. The Beta $\beta$ stands for the hyper-parameter in the Dirichlet distribution. As $\beta$ increases in [0.05, 0.07, 0.1, 0.3, 0.5, 1.0, 5.0], the data heterogeneity level reduces.
Illustrated in Figure~\ref{fig:hetero-level}, we can observe that (1) FedGC consistently outperforms these three algorithms in all different data heterogeneity levels. (2) As the heterogeneity level increases, the accuracy improvement brought by FedGC significantly elevates, which showcases the reliability of FedGC to mitigate heterogeneity, one of the intricate issues in FL.

\subsection{FedGC for Partial Client Participation Scenarios}
\label{app:partial_client}
Here, we conduct experiments of three baselines including FedAvg, FedProx, and SCAFFOLD on CIFAR-10 with Dirichlet distribution parameter $\beta$ = 0.1. Specifically, we set the communication round to 200, local iteration number to 100, and try different client number and participation rate. As illustrated in Table \ref{tab:partial_client}, we can observe that FedGC still significantly outperforms the baseline with no generated data under each circumstance. 

\begin{table}[t]
\setlength\tabcolsep{4.5pt}
\caption{Experiments of a scene in which only partial clients participate in training each round. We conduct experiments on three different total client numbers and several different participation rates. For example, client 200 and participation rate 5\% means randomly selecting 10 clients to participate in training each round. In each case, FedGC still significantly outperforms the baseline with no generative data. }
\label{tab:partial_client}
\begin{center}
\begin{tabular}{cc|ccc|cc|cc    }
\toprule
\multirow{2}{*}{Baseline} & Client & \multicolumn{3}{c|}{200} & \multicolumn{2}{c|}{100} & \multicolumn{2}{c}{50} \\
& Participation & 5\%  &  10\% & 20\% & 10\% & 20\% & 10\% & 20\% \\
\midrule
& Vanilla & 53.62 & 60.00 &  65.76 & 56.53 & 57.69 & 55.90 & 63.33   \\
\rowcolor{gray!15} \multirow{-2}{*}{\cellcolor{white}FedAvg} &+ FedGC & 68.93 & 74.06 & 75.74 & 74.16 & 74.26 & 75.34 & 77.20  \\ 
& Vanilla & 53.93 & 59.95 & 64.53 & 56.74 & 59.54 & 56.36 & 65.66  \\
\rowcolor{gray!15} \multirow{-2}{*}{\cellcolor{white}FedProx} &+ FedGC & 70.23 & 73.79  & 75.07 & 74.39 & 74.05 & 75.47 & 77.47   \\ 
 & Vanilla & 60.41 & 68.02 & 70.15 & 65.03 & 68.12 & 65.73 & 72.42  \\
\rowcolor{gray!15} \multirow{-2}{*}{\cellcolor{white}SCAFFOLD} &+ FedGC & 71.65 & 74.83 & 77.54 & 74.38 & 76.26 & 72.74 & 77.56   \\ 
\bottomrule
\end{tabular}
\end{center}
\end{table}

\subsection{Global-model-based Data Filtering}
\label{app:filter}
We propose global-model-based data filtering, where each client conducts data filtering on the client side according to the received global model before local model training.
Specifically, to determine which data to filter, a client feeds its generated data to the global model to evaluate the loss value for each data sample.
Then, each client selects the top $x \%$ data (we set $x=90$ here) and mixes the selected generated data with its real data.

Furthermore, since the global model might perform drastically differently on different categories, simply selecting according to the loss of all data samples may result in imbalanced filtering.
That is, this could make to global model filter out most of the samples where it performs poorly.
Addressing this, we further propose category-wise data filtering based on global model, which filers the same ratio of data for each category.

Here, we perform experiments on EuroSAT dataset with two heterogeneity levels in Table~\ref{tab:filter_1}. 
Vanilla denotes FedAvg itself, No F denotes FedGC without filtering, F@50 denotes filtering from round 50, F@50-C denotes category-wise filtering. 
From the table, we see that (1) under a high heterogeneity level, F@75 contributes to higher performance than No F, even with only 90\% of data at final rounds.
(2) Category-wise filtering generally performs better than unified filtering, indicating its effectiveness.
(3) Nevertheless, such filtering technique can not always ensure performance improvement, calling for more future work.
The performance drop could result from reduced number of data samples and ineffective filtering.

Overall, here we just provide an initial attempt to consider the potential of data filtering.
We believe more future works could be proposed to better filter the generated data such that we could use the generated data more efficiently.

\begin{table}[t!]
\caption{Experiments of global-model-based data filtering. We conduct our initial attempt on EuroSAT dataset with two heterogeneity types. F@50 means start filtering after 50 communication rounds and C means filtering by each class.}
\label{tab:filter_1}
\begin{center}
\begin{tabular}{c|cccccc}
\toprule
Heterogeneity Level &  Vanilla & No F & F@50  &  F@75 & F@50-C &  F@75-C \\
\midrule
High & 53.82 & 74.83 &72.96 & 74.93 & 73.50 &  74.20\\
Low & 75.59 & 84.46 & 83.82 & 83.83 & 84.19 & 83.83 \\
\bottomrule
\end{tabular}
\end{center}
\end{table}

\end{document}

%% file: math_commands.tex
\usepackage{amsmath,amsfonts,bm}

\def\eqref#1{equation~\ref{#1}}

\def\1{\bm{1}}

\def\vx{{\bm{x}}}

\def\vz{{\bm{z}}}

\DeclareMathAlphabet{\mathsfit}{\encodingdefault}{\sfdefault}{m}{sl}
\SetMathAlphabet{\mathsfit}{bold}{\encodingdefault}{\sfdefault}{bx}{n}













%% file: main.bbl
\begin{thebibliography}{50}
\providecommand{\natexlab}[1]{#1}
\providecommand{\url}[1]{\texttt{#1}}
\expandafter\ifx\csname urlstyle\endcsname\relax
  \providecommand{\doi}[1]{doi: #1}\else
  \providecommand{\doi}{doi: \begingroup \urlstyle{rm}\Url}\fi

\bibitem[Acar et~al.(2020)Acar, Zhao, Matas, Mattina, Whatmough, and Saligrama]{feddyn}
Durmus Alp~Emre Acar, Yue Zhao, Ramon Matas, Matthew Mattina, Paul Whatmough, and Venkatesh Saligrama.
\newblock Federated learning based on dynamic regularization.
\newblock In \emph{International Conference on Learning Representations}, 2020.

\bibitem[Brown et~al.(2020)Brown, Mann, Ryder, Subbiah, Kaplan, Dhariwal, Neelakantan, Shyam, Sastry, Askell, et~al.]{brown2020language}
Tom Brown, Benjamin Mann, Nick Ryder, Melanie Subbiah, Jared~D Kaplan, Prafulla Dhariwal, Arvind Neelakantan, Pranav Shyam, Girish Sastry, Amanda Askell, et~al.
\newblock Language models are few-shot learners.
\newblock \emph{Advances in neural information processing systems}, 33:\penalty0 1877--1901, 2020.

\bibitem[Caldas et~al.(2018)Caldas, Duddu, Wu, Li, Kone{\v{c}}n{\`y}, McMahan, Smith, and Talwalkar]{leaf}
Sebastian Caldas, Sai Meher~Karthik Duddu, Peter Wu, Tian Li, Jakub Kone{\v{c}}n{\`y}, H~Brendan McMahan, Virginia Smith, and Ameet Talwalkar.
\newblock Leaf: A benchmark for federated settings.
\newblock \emph{arXiv preprint arXiv:1812.01097}, 2018.

\bibitem[Celard et~al.(2023)Celard, Iglesias, Sorribes-Fdez, Romero, Vieira, and Borrajo]{celard2023survey}
Pedro Celard, EL~Iglesias, JM~Sorribes-Fdez, Rub{\'e}n Romero, A~Seara Vieira, and L~Borrajo.
\newblock A survey on deep learning applied to medical images: from simple artificial neural networks to generative models.
\newblock \emph{Neural Computing and Applications}, 35\penalty0 (3):\penalty0 2291--2323, 2023.

\bibitem[Chen et~al.(2022)Chen, Tu, Li, Shen, and Chao]{chen2022importance}
Hong-You Chen, Cheng-Hao Tu, Ziwei Li, Han~Wei Shen, and Wei-Lun Chao.
\newblock On the importance and applicability of pre-training for federated learning.
\newblock In \emph{The Eleventh International Conference on Learning Representations}, 2022.

\bibitem[Chen et~al.(2020)Chen, Kornblith, Norouzi, and Hinton]{chen2020simple}
Ting Chen, Simon Kornblith, Mohammad Norouzi, and Geoffrey Hinton.
\newblock A simple framework for contrastive learning of visual representations.
\newblock In \emph{International conference on machine learning}, pp.\  1597--1607. PMLR, 2020.

\bibitem[Eysenbach et~al.(2023)]{eysenbach2023role}
Gunther Eysenbach et~al.
\newblock The role of chatgpt, generative language models, and artificial intelligence in medical education: a conversation with chatgpt and a call for papers.
\newblock \emph{JMIR Medical Education}, 9\penalty0 (1):\penalty0 e46885, 2023.

\bibitem[Fang et~al.(2013)Fang, Xu, and Rockmore]{vlcs}
Chen Fang, Ye~Xu, and Daniel~N Rockmore.
\newblock Unbiased metric learning: On the utilization of multiple datasets and web images for softening bias.
\newblock In \emph{Proceedings of the IEEE International Conference on Computer Vision}, pp.\  1657--1664, 2013.

\bibitem[Fix \& Hodges(1989)Fix and Hodges]{fix1989discriminatory}
Evelyn Fix and Joseph~Lawson Hodges.
\newblock Discriminatory analysis. nonparametric discrimination: Consistency properties.
\newblock \emph{International Statistical Review/Revue Internationale de Statistique}, 57\penalty0 (3):\penalty0 238--247, 1989.

\bibitem[He et~al.(2016)He, Zhang, Ren, and Sun]{resnet}
Kaiming He, Xiangyu Zhang, Shaoqing Ren, and Jian Sun.
\newblock Deep residual learning for image recognition.
\newblock In \emph{Proceedings of the IEEE conference on computer vision and pattern recognition}, pp.\  770--778, 2016.

\bibitem[He et~al.(2022)He, Sun, Yu, Xue, Zhang, Torr, Bai, and QI]{he2022synthetic}
Ruifei He, Shuyang Sun, Xin Yu, Chuhui Xue, Wenqing Zhang, Philip Torr, Song Bai, and XIAOJUAN QI.
\newblock Is synthetic data from generative models ready for image recognition?
\newblock In \emph{The Eleventh International Conference on Learning Representations}, 2022.

\bibitem[Helber et~al.(2019)Helber, Bischke, Dengel, and Borth]{helber2019eurosat}
Patrick Helber, Benjamin Bischke, Andreas Dengel, and Damian Borth.
\newblock Eurosat: A novel dataset and deep learning benchmark for land use and land cover classification.
\newblock \emph{IEEE Journal of Selected Topics in Applied Earth Observations and Remote Sensing}, 12\penalty0 (7):\penalty0 2217--2226, 2019.

\bibitem[Hsu et~al.(2019)Hsu, Qi, and Brown]{fedavgm}
Tzu-Ming~Harry Hsu, Hang Qi, and Matthew Brown.
\newblock Measuring the effects of non-identical data distribution for federated visual classification.
\newblock \emph{arXiv preprint arXiv:1909.06335}, 2019.

\bibitem[Hu et~al.(2021)Hu, Wallis, Allen-Zhu, Li, Wang, Wang, Chen, et~al.]{hu2021lora}
Edward~J Hu, Phillip Wallis, Zeyuan Allen-Zhu, Yuanzhi Li, Shean Wang, Lu~Wang, Weizhu Chen, et~al.
\newblock Lora: Low-rank adaptation of large language models.
\newblock In \emph{International Conference on Learning Representations}, 2021.

\bibitem[Jhunjhunwala et~al.(2022)Jhunjhunwala, Wang, and Joshi]{fedexp}
Divyansh Jhunjhunwala, Shiqiang Wang, and Gauri Joshi.
\newblock Fedexp: Speeding up federated averaging via extrapolation.
\newblock In \emph{The Eleventh International Conference on Learning Representations}, 2022.

\bibitem[Kairouz et~al.(2021)Kairouz, McMahan, Avent, Bellet, Bennis, Bhagoji, Bonawitz, Charles, Cormode, Cummings, et~al.]{advances}
Peter Kairouz, H~Brendan McMahan, Brendan Avent, Aur{\'e}lien Bellet, Mehdi Bennis, Arjun~Nitin Bhagoji, Kallista Bonawitz, Zachary Charles, Graham Cormode, Rachel Cummings, et~al.
\newblock Advances and open problems in federated learning.
\newblock \emph{Foundations and Trends{\textregistered} in Machine Learning}, 14\penalty0 (1--2):\penalty0 1--210, 2021.

\bibitem[Karimireddy et~al.(2020)Karimireddy, Kale, Mohri, Reddi, Stich, and Suresh]{scaffold}
Sai~Praneeth Karimireddy, Satyen Kale, Mehryar Mohri, Sashank Reddi, Sebastian Stich, and Ananda~Theertha Suresh.
\newblock Scaffold: Stochastic controlled averaging for federated learning.
\newblock In \emph{International Conference on Machine Learning}, pp.\  5132--5143. PMLR, 2020.

\bibitem[Kazerouni et~al.(2022)Kazerouni, Aghdam, Heidari, Azad, Fayyaz, Hacihaliloglu, and Merhof]{kazerouni2022diffusion}
Amirhossein Kazerouni, Ehsan~Khodapanah Aghdam, Moein Heidari, Reza Azad, Mohsen Fayyaz, Ilker Hacihaliloglu, and Dorit Merhof.
\newblock Diffusion models for medical image analysis: A comprehensive survey.
\newblock \emph{arXiv preprint arXiv:2211.07804}, 2022.

\bibitem[Krizhevsky et~al.(2009)Krizhevsky, Hinton, et~al.]{cifar10}
Alex Krizhevsky, Geoffrey Hinton, et~al.
\newblock Learning multiple layers of features from tiny images.
\newblock 2009.

\bibitem[Li et~al.(2021)Li, He, and Song]{moon}
Qinbin Li, Bingsheng He, and Dawn Song.
\newblock Model-contrastive federated learning.
\newblock In \emph{Proceedings of the IEEE/CVF Conference on Computer Vision and Pattern Recognition}, pp.\  10713--10722, 2021.

\bibitem[Li et~al.(2020{\natexlab{a}})Li, Sahu, Talwalkar, and Smith]{li_survey}
Tian Li, Anit~Kumar Sahu, Ameet Talwalkar, and Virginia Smith.
\newblock Federated learning: Challenges, methods, and future directions.
\newblock \emph{IEEE Signal Processing Magazine}, 37\penalty0 (3):\penalty0 50--60, 2020{\natexlab{a}}.

\bibitem[Li et~al.(2020{\natexlab{b}})Li, Sahu, Zaheer, Sanjabi, Talwalkar, and Smith]{fedprox}
Tian Li, Anit~Kumar Sahu, Manzil Zaheer, Maziar Sanjabi, Ameet Talwalkar, and Virginia Smith.
\newblock Federated optimization in heterogeneous networks.
\newblock \emph{Proceedings of Machine Learning and Systems}, 2:\penalty0 429--450, 2020{\natexlab{b}}.

\bibitem[Li et~al.(2019)Li, Huang, Yang, Wang, and Zhang]{li2019convergence}
Xiang Li, Kaixuan Huang, Wenhao Yang, Shusen Wang, and Zhihua Zhang.
\newblock On the convergence of fedavg on non-iid data.
\newblock In \emph{International Conference on Learning Representations}, 2019.

\bibitem[McMahan et~al.(2017)McMahan, Moore, Ramage, Hampson, and y~Arcas]{fedavg}
Brendan McMahan, Eider Moore, Daniel Ramage, Seth Hampson, and Blaise~Aguera y~Arcas.
\newblock Communication-efficient learning of deep networks from decentralized data.
\newblock In \emph{Artificial intelligence and statistics}, pp.\  1273--1282. PMLR, 2017.

\bibitem[Nguyen et~al.(2022)Nguyen, Wang, Malik, Sanjabi, and Rabbat]{nguyen2022begin}
John Nguyen, Jianyu Wang, Kshitiz Malik, Maziar Sanjabi, and Michael Rabbat.
\newblock Where to begin? on the impact of pre-training and initialization in federated learning.
\newblock In \emph{The Eleventh International Conference on Learning Representations}, 2022.

\bibitem[Nichol et~al.(2022)Nichol, Dhariwal, Ramesh, Shyam, Mishkin, Mcgrew, Sutskever, and Chen]{nichol2022glide}
Alexander~Quinn Nichol, Prafulla Dhariwal, Aditya Ramesh, Pranav Shyam, Pamela Mishkin, Bob Mcgrew, Ilya Sutskever, and Mark Chen.
\newblock Glide: Towards photorealistic image generation and editing with text-guided diffusion models.
\newblock In \emph{International Conference on Machine Learning}, pp.\  16784--16804. PMLR, 2022.

\bibitem[OpenAI(2023)]{openai2023gpt4}
OpenAI.
\newblock Gpt-4 technical report, 2023.

\bibitem[Ouyang et~al.(2022)Ouyang, Wu, Jiang, Almeida, Wainwright, Mishkin, Zhang, Agarwal, Slama, Ray, et~al.]{ouyang2022training}
Long Ouyang, Jeffrey Wu, Xu~Jiang, Diogo Almeida, Carroll Wainwright, Pamela Mishkin, Chong Zhang, Sandhini Agarwal, Katarina Slama, Alex Ray, et~al.
\newblock Training language models to follow instructions with human feedback.
\newblock \emph{Advances in Neural Information Processing Systems}, 35:\penalty0 27730--27744, 2022.

\bibitem[Proakis(2008)]{proakis2008digital}
John~G Proakis.
\newblock \emph{Digital communications}.
\newblock McGraw-Hill, Higher Education, 2008.

\bibitem[Radford et~al.(2019)Radford, Wu, Child, Luan, Amodei, Sutskever, et~al.]{radford2019language}
Alec Radford, Jeffrey Wu, Rewon Child, David Luan, Dario Amodei, Ilya Sutskever, et~al.
\newblock Language models are unsupervised multitask learners.
\newblock \emph{OpenAI blog}, 1\penalty0 (8):\penalty0 9, 2019.

\bibitem[Radford et~al.(2021)Radford, Kim, Hallacy, Ramesh, Goh, Agarwal, Sastry, Askell, Mishkin, Clark, et~al.]{radford2021learning}
Alec Radford, Jong~Wook Kim, Chris Hallacy, Aditya Ramesh, Gabriel Goh, Sandhini Agarwal, Girish Sastry, Amanda Askell, Pamela Mishkin, Jack Clark, et~al.
\newblock Learning transferable visual models from natural language supervision.
\newblock In \emph{International conference on machine learning}, pp.\  8748--8763. PMLR, 2021.

\bibitem[Reddi et~al.(2020)Reddi, Charles, Zaheer, Garrett, Rush, Kone{\v{c}}n{\`y}, Kumar, and McMahan]{fedopt}
Sashank~J Reddi, Zachary Charles, Manzil Zaheer, Zachary Garrett, Keith Rush, Jakub Kone{\v{c}}n{\`y}, Sanjiv Kumar, and Hugh~Brendan McMahan.
\newblock Adaptive federated optimization.
\newblock In \emph{International Conference on Learning Representations}, 2020.

\bibitem[Rombach et~al.(2022)Rombach, Blattmann, Lorenz, Esser, and Ommer]{rombach2022high}
Robin Rombach, Andreas Blattmann, Dominik Lorenz, Patrick Esser, and Bj{\"o}rn Ommer.
\newblock High-resolution image synthesis with latent diffusion models.
\newblock In \emph{Proceedings of the IEEE/CVF Conference on Computer Vision and Pattern Recognition}, pp.\  10684--10695, 2022.

\bibitem[Ronneberger et~al.(2015)Ronneberger, Fischer, and Brox]{u-net}
Olaf Ronneberger, Philipp Fischer, and Thomas Brox.
\newblock U-net: Convolutional networks for biomedical image segmentation.
\newblock In \emph{Medical Image Computing and Computer-Assisted Intervention--MICCAI 2015: 18th International Conference, Munich, Germany, October 5-9, 2015, Proceedings, Part III 18}, pp.\  234--241. Springer, 2015.

\bibitem[Sablayrolles et~al.(2019)Sablayrolles, Douze, Schmid, Ollivier, and J{\'e}gou]{sablayrolles2019white}
Alexandre Sablayrolles, Matthijs Douze, Cordelia Schmid, Yann Ollivier, and Herv{\'e} J{\'e}gou.
\newblock White-box vs black-box: Bayes optimal strategies for membership inference.
\newblock In \emph{International Conference on Machine Learning}, pp.\  5558--5567. PMLR, 2019.

\bibitem[Saharia et~al.(2022)Saharia, Chan, Saxena, Li, Whang, Denton, Ghasemipour, Gontijo~Lopes, Karagol~Ayan, Salimans, et~al.]{saharia2022photorealistic}
Chitwan Saharia, William Chan, Saurabh Saxena, Lala Li, Jay Whang, Emily~L Denton, Kamyar Ghasemipour, Raphael Gontijo~Lopes, Burcu Karagol~Ayan, Tim Salimans, et~al.
\newblock Photorealistic text-to-image diffusion models with deep language understanding.
\newblock \emph{Advances in Neural Information Processing Systems}, 35:\penalty0 36479--36494, 2022.

\bibitem[Shi et~al.(2022)Shi, Liang, Zhang, Tan, and Bai]{feddecorr}
Yujun Shi, Jian Liang, Wenqing Zhang, Vincent~YF Tan, and Song Bai.
\newblock Towards understanding and mitigating dimensional collapse in heterogeneous federated learning.
\newblock \emph{arXiv preprint arXiv:2210.00226}, 2022.

\bibitem[Shipard et~al.(2023)Shipard, Wiliem, Thanh, Xiang, and Fookes]{shipard2023diversity}
Jordan Shipard, Arnold Wiliem, Kien~Nguyen Thanh, Wei Xiang, and Clinton Fookes.
\newblock Diversity is definitely needed: Improving model-agnostic zero-shot classification via stable diffusion.
\newblock In \emph{Proceedings of the IEEE/CVF Conference on Computer Vision and Pattern Recognition}, pp.\  769--778, 2023.

\bibitem[Touvron et~al.(2023)Touvron, Martin, Stone, Albert, Almahairi, Babaei, Bashlykov, Batra, Bhargava, Bhosale, et~al.]{llama2}
Hugo Touvron, Louis Martin, Kevin Stone, Peter Albert, Amjad Almahairi, Yasmine Babaei, Nikolay Bashlykov, Soumya Batra, Prajjwal Bhargava, Shruti Bhosale, et~al.
\newblock Llama 2: Open foundation and fine-tuned chat models.
\newblock \emph{arXiv preprint arXiv:2307.09288}, 2023.

\bibitem[Tschandl et~al.(2018)Tschandl, Rosendahl, and Kittler]{tschandl2018ham10000}
Philipp Tschandl, Cliff Rosendahl, and Harald Kittler.
\newblock The ham10000 dataset, a large collection of multi-source dermatoscopic images of common pigmented skin lesions.
\newblock \emph{Scientific data}, 5\penalty0 (1):\penalty0 1--9, 2018.

\bibitem[Wang et~al.(2020{\natexlab{a}})Wang, Yurochkin, Sun, Papailiopoulos, and Khazaeni]{fedma}
Hongyi Wang, Mikhail Yurochkin, Yuekai Sun, Dimitris Papailiopoulos, and Yasaman Khazaeni.
\newblock Federated learning with matched averaging.
\newblock In \emph{International Conference on Learning Representations}, 2020{\natexlab{a}}.
\newblock URL \url{https://openreview.net/forum?id=BkluqlSFDS}.

\bibitem[Wang et~al.(2020{\natexlab{b}})Wang, Liu, Liang, Joshi, and Poor]{fednova}
Jianyu Wang, Qinghua Liu, Hao Liang, Gauri Joshi, and H~Vincent Poor.
\newblock Tackling the objective inconsistency problem in heterogeneous federated optimization.
\newblock \emph{Advances in neural information processing systems}, 33:\penalty0 7611--7623, 2020{\natexlab{b}}.

\bibitem[Wang et~al.(2021)Wang, Charles, Xu, Joshi, McMahan, Al-Shedivat, Andrew, Avestimehr, Daly, Data, et~al.]{wangsurvey}
Jianyu Wang, Zachary Charles, Zheng Xu, Gauri Joshi, H~Brendan McMahan, Maruan Al-Shedivat, Galen Andrew, Salman Avestimehr, Katharine Daly, Deepesh Data, et~al.
\newblock A field guide to federated optimization.
\newblock \emph{arXiv preprint arXiv:2107.06917}, 2021.

\bibitem[Yang et~al.(2019)Yang, Liu, Chen, and Tong]{yang_survey}
Qiang Yang, Yang Liu, Tianjian Chen, and Yongxin Tong.
\newblock Federated machine learning: Concept and applications.
\newblock \emph{ACM Transactions on Intelligent Systems and Technology (TIST)}, 10\penalty0 (2):\penalty0 1--19, 2019.

\bibitem[Ye et~al.(2022)Ye, Ni, Xu, Wang, Chen, and Eldar]{fedfm}
Rui Ye, Zhenyang Ni, Chenxin Xu, Jianyu Wang, Siheng Chen, and Yonina~C Eldar.
\newblock Fedfm: Anchor-based feature matching for data heterogeneity in federated learning.
\newblock \emph{arXiv preprint arXiv:2210.07615}, 2022.

\bibitem[Ye et~al.(2023)Ye, Xu, Wang, Xu, Chen, and Wang]{feddisco}
Rui Ye, Mingkai Xu, Jianyu Wang, Chenxin Xu, Siheng Chen, and Yanfeng Wang.
\newblock Feddisco: Federated learning with discrepancy-aware collaboration.
\newblock \emph{arXiv preprint arXiv:2305.19229}, 2023.

\bibitem[Yu et~al.(2021)Yu, Zhang, Chen, Yin, and Liu]{yu2021does}
Da~Yu, Huishuai Zhang, Wei Chen, Jian Yin, and Tie-Yan Liu.
\newblock How does data augmentation affect privacy in machine learning?
\newblock In \emph{Proceedings of the AAAI Conference on Artificial Intelligence}, volume~35, pp.\  10746--10753, 2021.

\bibitem[Zhang et~al.(2015)Zhang, Zhao, and LeCun]{yahoo}
Xiang Zhang, Junbo Zhao, and Yann LeCun.
\newblock Character-level convolutional networks for text classification.
\newblock \emph{Advances in neural information processing systems}, 28, 2015.

\bibitem[Zhang et~al.(2022)Zhang, Kang, Chen, Fan, and Yang]{no_free_lunch}
Xiaojin Zhang, Yan Kang, Kai Chen, Lixin Fan, and Qiang Yang.
\newblock Trading off privacy, utility and efficiency in federated learning.
\newblock \emph{arXiv preprint arXiv:2209.00230}, 2022.

\bibitem[Zhou et~al.(2020)Zhou, Yang, Hospedales, and Xiang]{pacs}
Kaiyang Zhou, Yongxin Yang, Timothy Hospedales, and Tao Xiang.
\newblock Deep domain-adversarial image generation for domain generalisation.
\newblock In \emph{Proceedings of the AAAI conference on artificial intelligence}, volume~34, pp.\  13025--13032, 2020.

\end{thebibliography}
